\documentclass[journal,twoside]{IEEEtran}
\usepackage{cite}
\usepackage{url}
\usepackage{ragged2e}
\usepackage{epsfig}
\usepackage{graphicx}
\usepackage{amsmath,amssymb} 
\usepackage{subfigure}
\usepackage{algorithm}
\usepackage{algorithmicx}
\usepackage{algpseudocode}
\usepackage{graphics}
\usepackage{threeparttable}
\usepackage{color}
\usepackage[normalem]{ulem}
\usepackage{multirow}
\usepackage{float}
\usepackage{amsfonts}
\usepackage{bm}
\usepackage{array}
\usepackage[table]{xcolor}
\usepackage{colortbl}
\usepackage{pifont}
\usepackage{diagbox}
\usepackage{rotating}
\usepackage{booktabs}
\usepackage{overpic}
\usepackage{makecell}
\usepackage{textcomp}
\usepackage{contour}
\usepackage{enumitem}
\usepackage[colorlinks,linkcolor=blue,anchorcolor=blue,citecolor=blue,]{hyperref}

\RequirePackage{silence}
\definecolor{mygray}{gray}{.92}

\newcommand{\tabref}[1]{Tab. \ref{#1}}

\newcommand{\figref}[1]{Fig. \ref{#1}}
\newcommand{\supp}[1]{\textcolor{magenta}{#1}}

\newcommand{\secref}[1]{$\S~$\ref{#1}}
\newcommand{\sArt}{state-of-the-art }

\def\ie{\emph{i.e.}}
\def\eg{\emph{e.g.}}

\def\etal{{\em et al.~}}

\def\ourmodel{\emph{CalibNet}}

\definecolor{mygray}{gray}{.92}

\graphicspath{{./Imgs/}}
\DeclareGraphicsExtensions{.jpg,.pdf,.png}

\ifCLASSINFOpdf
\else
\fi
\begin{document}
%
\title{CalibNet: Dual-branch Cross-modal Calibration for RGB-D Salient Instance Segmentation}

%
%
%
%
\author{Jialun Pei, \IEEEmembership{Member, IEEE}, Tao Jiang, He Tang, Nian Liu, \IEEEmembership{Member, IEEE}, Yueming Jin, \IEEEmembership{Member, IEEE}, \\Deng-Ping Fan, \IEEEmembership{Senior Member, IEEE}, Pheng-Ann Heng, \IEEEmembership{Senior Member, IEEE}%
	\IEEEcompsocitemizethanks{
		\IEEEcompsocthanksitem 
		Jialun Pei, Pheng-Ann Heng are with the School of Computer Science and Engineering, The Chinese University of Hong Kong, HKSAR, China (e-mail: jialunpei@cuhk.edu.hk; pheng@cse.cuhk.edu.hk).
		\IEEEcompsocthanksitem 
		Tao Jiang, He Tang are with the School of Software Engineering, Huazhong University of Science and Technology, Wuhan, China (e-mail: jtao99@hust.edu.cn; hetang@hust.edu.cn).
		\IEEEcompsocthanksitem 
		Nian Liu is with the Computer Vision Department, Mohamed Bin Zayed University of Artificial Intelligence, Abu Dhabi, United Arab Emirates
(e-mail: liunian228@gmail.com).
		\IEEEcompsocthanksitem 
		Yueming Jin is with the Department of Electrical and Computer Engineering, National University of Singapore, Singapore (e-mail: ymjin@nus.edu.sg). 
		\IEEEcompsocthanksitem 
             Deng-Ping Fan is with Nankai International Advanced Research Institute (Shenzhen Futian), Nankai University, Shenzhen, China and he is also with College of Computer Science, Nankai University, Tianjin, China (e-mail: dengpfan@gmail.com).
             \IEEEcompsocthanksitem 
		   Corresponding author: Deng-Ping Fan.
	}
}

%
%

\markboth{IEEE TRANSACTIONS ON IMAGE PROCESSING}{Pei \MakeLowercase{\textit{et al.}}: CalibNet: Dual-branch Cross-modal Calibration for RGB-D Salient Instance Segmentation}
%



\maketitle

\begin{abstract}
\justifying
In this study, we propose a novel approach for RGB-D salient instance segmentation using a dual-branch cross-modal feature calibration architecture called \textbf{\ourmodel}. 
Our method simultaneously calibrates depth and RGB features in the kernel and mask branches to generate instance-aware kernels and mask features. \ourmodel~consists of three simple modules, a dynamic interactive kernel (DIK) and a weight-sharing fusion (WSF), which work together to generate effective instance-aware kernels and integrate cross-modal features. To improve the quality of depth features, we incorporate a depth similarity assessment (DSA) module prior to DIK and WSF.
In addition, we further contribute a new \textbf{DSIS} dataset, which contains 1,940 images with elaborate instance-level annotations.
Extensive experiments on three challenging benchmarks show that~\ourmodel~yields a promising result, \ie, 58.0\% AP with 320$\times$480 input size on the COME15K-E test set, which significantly surpasses the alternative frameworks. 
Our code and dataset will be publicly available at: \supp{\url{https://github.com/PJLallen/CalibNet}.}
\end{abstract}

\begin{IEEEkeywords}
RGB-D salient instance segmentation, cross-modal fusion, instance-level segmentation, transformer.
\end{IEEEkeywords}

%
\IEEEpeerreviewmaketitle

\section{Introduction}

\begin{figure}[t!]
\centering
    \begin{overpic}[width=\linewidth]{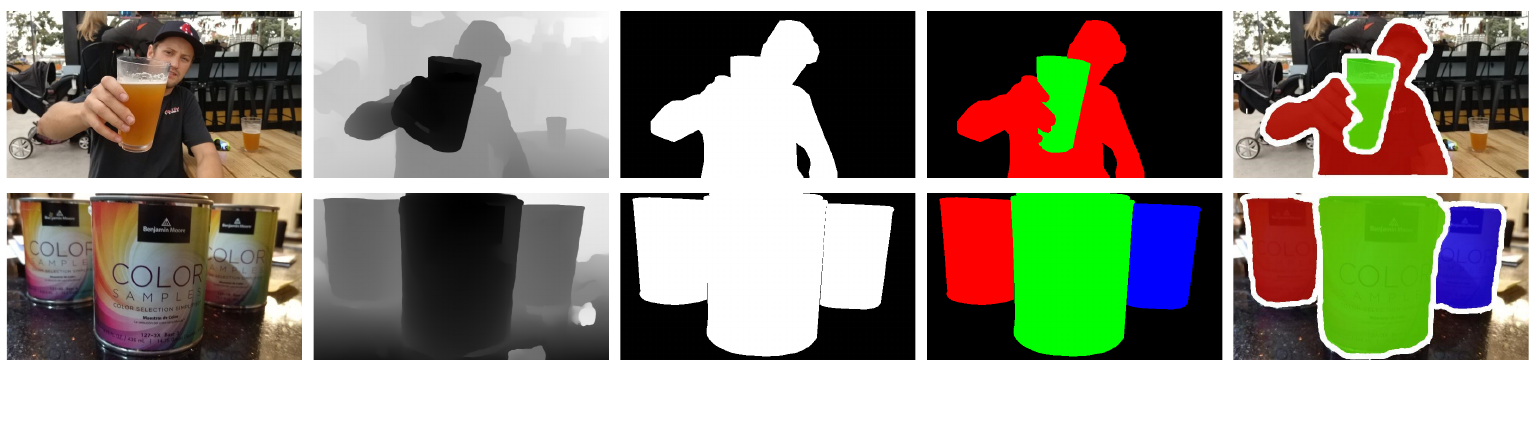}
        \put(6,0.7){\small{RGB}}
        \put(25,0.7){\small{Depth}}
        \put(42.5,0.7){\small{Object-GT}}
        \put(61.5,0.7){\small{Instance-GT}}
        \put(86.5,0.7){\small{Ours}}
    \end{overpic}
	\caption{Illustration of the RGB-D salient instance segmentation task with the proposed~\ourmodel~predictions. Our method propels RGB-D saliency detection to instance-level identification.}\label{DSISsample}
\end{figure}

%
%
%
%
\IEEEPARstart{I}{n} recent years, depth maps have been introduced as an additional source of information to provide intuitive cues of spatial structure in the RGB-D Salient Object Detection (SOD) task~\cite{zhou2021rgb,liu2021visual}. 
While most RGB-D SOD approaches~\cite{ji2021calibrated,zhou2021specificity,wen2021dynamic} have effectively utilized depth information by developing cross-modal feature fusion modules and depth quality assessment strategies in the backbone or decoder structure, relatively few approaches have incorporated depth awareness for instance-level saliency detection.
Compared to SOD, salient instance segmentation (SIS) offers an opportunity to determine the location and number of each instance, which is particularly valuable for practical applications such as autonomous driving~\cite{chen2021multisiam}, video surveillance~\cite{feng2021mist}, and scene understanding~\cite{jin2022exploring}.
Therefore, our work aims to take the first step towards incorporating depth-aware cues to enhance the RGB feature representation for RGB-D SIS (see \figref{DSISsample}). 

\begin{figure*}[t!]
\centering
    \begin{overpic}[width=\textwidth]{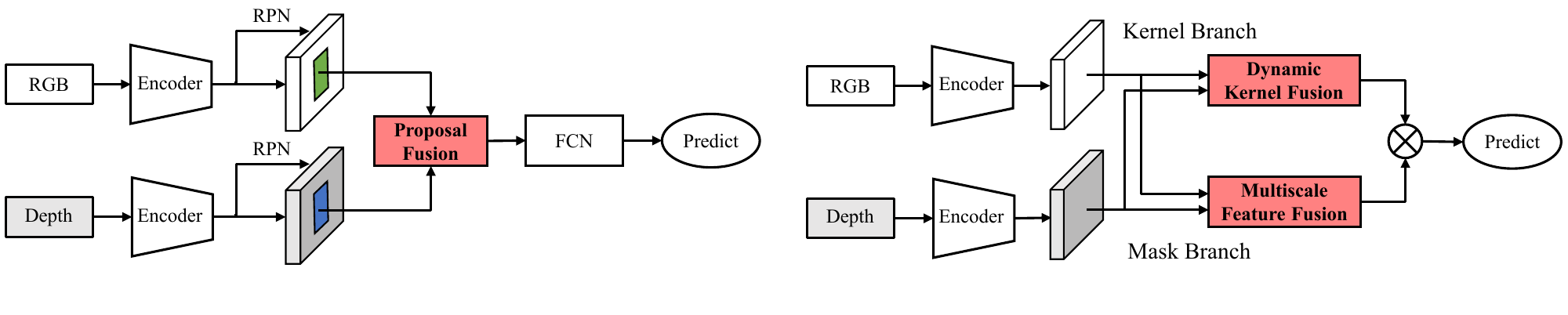}
        \put(16,0.2){\small{(a) Proposal Fusion}}
        \put(64,0.2){\small{(b) Dual-branch Fusion (Ours)}}
    \end{overpic}
 \caption{Comparison of two kinds of fusion architectures for RGB-D instance-level segmentation. (a) Proposal fusion in a two-stage manner~\cite{xu2020outdoor}; (b) Our dual-branch fusion in a one-stage manner.}\label{fusionstyle}
\end{figure*}

Currently, there are two typical RGB instance-level segmentation frameworks: the RPN-based framework (\ie, two-stage scheme)~\cite{he2017mask, fan2019s4net} and the class\&mask-based framework (\ie, one-stage scheme)~\cite{wang2021solo,cheng2022masked}. 
The two-stage RGB-D scheme (\figref{fusionstyle} (a)) requires additional proposal generation and refinement stages, which increase the computational complexity and time cost. 
Hence, it is desirable to exploit one-stage architectures to efficiently fuse multiscale RGB and depth features and utilize cross-modal features to share refined mutual information for generating instance-aware embeddings. However, one-stage RGB-D schemes may struggle with cases where depth and RGB information are inconsistent or noisy, potentially leading to inaccurate instance segmentation results.

Based on the above observations, we propose a dual-branch cross-modal calibration network, called \textbf{\ourmodel}, to identify salient instances in a one-stage manner. 
As shown in~\figref{fusionstyle} (b), compared to proposal fusion in two-stage RGB-D instance-level models, the proposed \ourmodel~employs a bilateral fusion strategy to integrate depth information in both the kernel and mask branches. 
The essence of the dual-branch cross-modal calibration lies in the efficient exploitation of depth cues to enhance multiscale RGB features in both branches, ensuring strong consistency between the instance-aware kernels and the generated mask features.
To this end, in the kernel branch, we introduce a Dynamic Interactive Kernel (\textbf{DIK}) module to produce informative and structural attention maps using RGB and depth features, followed by combining global features for dynamic kernel generation. 
In the mask branch, the main challenge is the suppression of low-quality structure information from depth features and the efficient fusion of valuable spatial cues with RGB features.
To bridge this gap, a weight-sharing fusion (\textbf{WSF}) is proposed to efficiently exploit cross-modal mutual information and integrate multiscale shared features to enhance mask feature representation. 
Furthermore, a Depth Similarity Assessment (\textbf{DSA}) module is embedded to minimize noisy depth information and identify valuable complements. 
Overall, our approach aims to enhance the accuracy and reliability of the depth information used in the kernel and mask branches, resulting in precise and robust segmentation results. 
Additionally, we utilize bipartite matching~\cite{carion2020end} to facilitate label assignment and reduce redundant predictions, rather than relying on non-maximum suppression (NMS) post-processing. 
This enables us to achieve better segmentation performance while reducing computational overhead.

In addition, there are only two datasets available for RGB-D SIS task: COME15K~\cite{zhang2021rgb} and SIP~\cite{fan2020rethinking}. 
COME15K has the largest number of training samples but deficient annotation quality. SIP has highly accurate annotations, but only includes 929 images and focuses solely on the ``person" category. 
To promote the development of RGB-D SIS and better evaluate model generalization performance, it is crucial to construct a moderately scale, high-quality annotated test set that includes multiple categories of salient instances. 
Thus, we contribute a new dataset, namely \textbf{DSIS}, which includes 1,940 images with detailed instance-level labels as well as depth maps. To better understand the development of existing SOTA techniques, we further evaluate~\ourmodel~on three RGB-D SIS datasets, \ie, COME15K, SIP, and our DSIS, making it the first systematic investigation in the RGB-D SIS field. 

Our main contributions include:
%


%
\begin{itemize}
\item To our knowledge, \textbf{\ourmodel} is the pioneering framework for RGB-D SIS. The one-stage framework exploits depth information to optimize and interact with RGB features in both kernel and mask branches, showing strong performance on all three RGB-D SIS datasets.

\item Our framework contains a dynamic interactive kernel (DIK) module for highlighting spatially critical locations and dynamically generating instance-level kernels. Additionally, a weight-sharing fusion (WSF) module is introduced to improve the fusion efficiency of multiscale depth and RGB features. To calibrate depth features, a depth similarity assessment (DSA) module excavates valuable depth information.

\item We created a new dataset called \textbf{DSIS} to assess the generalization performance of RGB-D SIS methods. Using DSIS and two existing datasets, we conducted the first large-scale RGB-D SIS study, evaluating 19 different baselines to develop the RGB-D SIS community.
\end{itemize}

\section{Related Work}

\subsection{RGB-D Salient Object Detection}

The depth modality with rich geometrical information plays a vital role in saliency detection, particularly in more challenging situations. 
Benefiting from the powerful feature representation capability of CNNs, previous RGB-D SOD methods~\cite{qu2017rgbd,zhang2021bts,cong2017co,chen2020dpanet,zhang2021uncertainty} leverage multi-level RGB and depth features to attain favorable results.
These models~\cite{zhang2021bilateral,chen2020progressively,li2020icnet} mainly design cross-modal fusion strategies based on fully convolutional neural networks (FCNs)~\cite{long2015fully}, which can be grouped into three types: early fusion~\cite{qu2017rgbd,song2017depth,zhao2020single}, late fusion~\cite{desingh2013depth,han2017cnns,piao2020a2dele}, and middle multi-level fusion~\cite{chen2018progressively,zhao2019contrast,liu2020learning,zhang2020select,ji2020accurate,fan2020bbs,li2020cross,piao2019depth}. 
Early fusion concatenates the RGB image and depth map before feeding them into the encoder for unified processing. 
In contrast, the late-fusion strategy processes each modal independently and then fuses them in the last layer. 
To strengthen the connection between RGB and depth features, recent RGB-D SOD models frequently adopt middle multi-level fusion, which progressively interacts with cross-modal features at various scales and levels. 
%
%
For instance, Fu \etal~\cite{fu2021siamese} utilized a siamese architecture with a joint learning module and a cooperative fusion module to extract hierarchical features and exploit cross-modal complementarity.
Wu \etal~\cite{wu2021mobilesal} created a highly efficient network to improve feature representations and fuse RGB and depth features without incurring high computational costs. 
Later, Liu \etal~\cite{liu2021visual} developed a pure transformer model to extract global features in a sequence-to-sequence manner and fuse RGB and depth features via cross-attention. 
However, previous works have not utilized depth data to enhance feature representation in instance-level SOD models. Motivated by this observation, we strive to develop a novel cross-modal fusion framework for RGB-D SIS. 

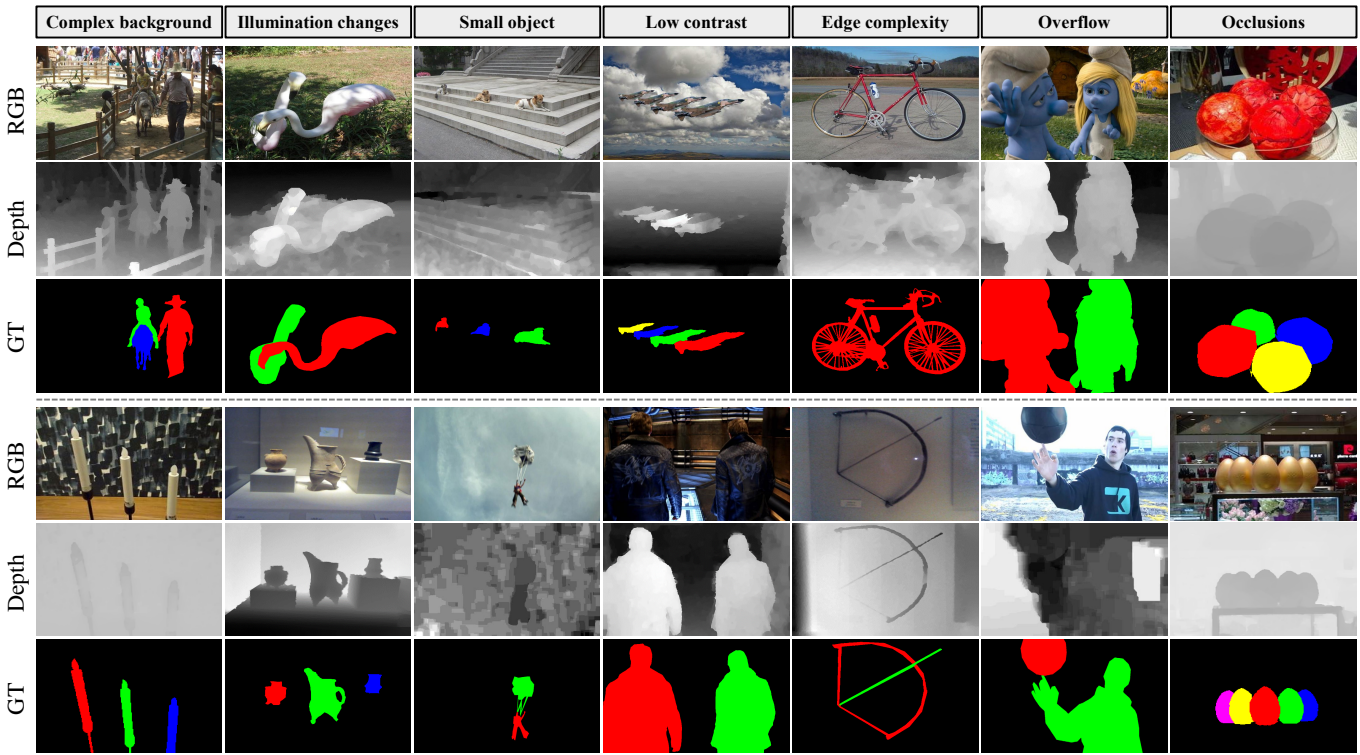
\begin{figure*}[t!]
	\centering
    \begin{overpic}[width=\textwidth]{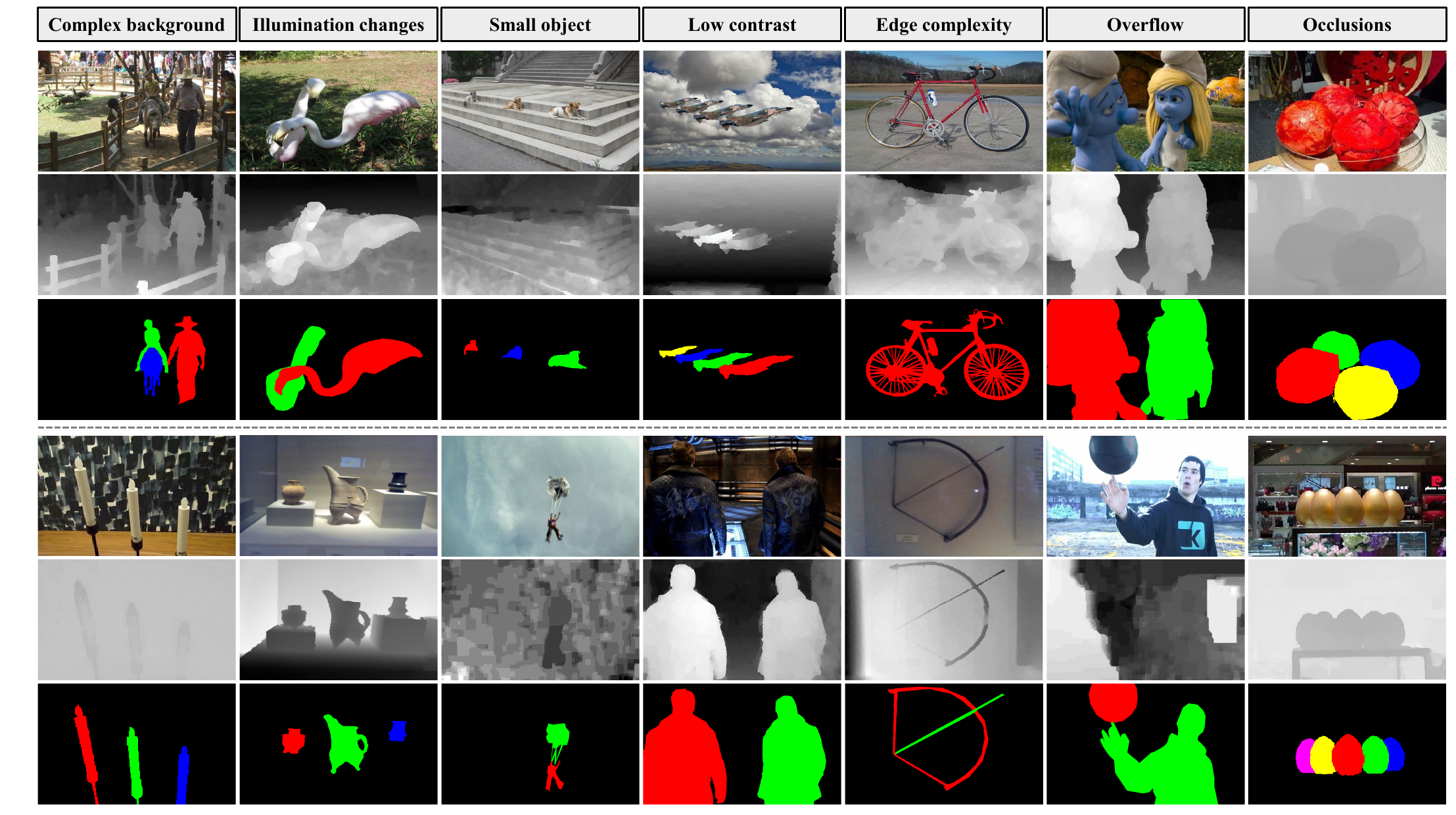}
        \put(0.5,46){\rotatebox{90}{\small{RGB}}}
        \put(0.5,37){\rotatebox{90}{\small{Depth}}}
        \put(0.5,30){\rotatebox{90}{\small{GT}}}
        \put(0.5,20){\rotatebox{90}{\small{RGB}}}
        \put(0.5,11){\rotatebox{90}{\small{Depth}}}
        \put(0.5,3.5){\rotatebox{90}{\small{GT}}}
    \end{overpic}
	\caption{Example of a diverse annotation of the proposed DSIS dataset.}\label{DSIS-show}
\end{figure*}

\subsection{Salient Instance Segmentation} 

SIS aims to identify individual instances within the salient region. 
It is critical to determine the number of salient instances together with the pixel-level masks.
Unlike the generic instance segmentation (GIS) task~\cite{he2017mask,wang2021solo,cheng2022sparse}, SIS is class-agnostic because the identification of salient instances is based on the salient region, rather than specific class annotations. 
Li \etal~\cite{li2017instance} first proposed the SIS task and developed a multi-stage model, along with the first SIS dataset (ILSO). Since then, several SIS frameworks have been proposed, mostly based on the Mask R-CNN~\cite{he2017mask}.
S4Net~\cite{fan2019s4net} introduced the first single-stage end-to-end framework for segmenting salient instances.
On the basis of this, RDPNet~\cite{wu2021regularized} adopted a regularized dense pyramid network to improve the feature pyramid and suppress non-informative regions. 
Similarly, SCG~\cite{liu2021scg} added the saliency and contour branches in Mask R-CNN to enhance the recognition of salient instances.
Unlike fully-supervised learning, Tian \etal~\cite{tian2022learning} proposed a weakly-supervised network that adopts subitizing and class labels as weak supervision for SIS. 
More recently, OQTR~\cite{pei2022transformer} brought a transformer-based model that contains a cross-fusion module and salient queries to efficiently merge global features.  
Although these methods have led to the rapid development of the SIS task, they only focus on a single modality. 

\subsection{Instance Segmentation with RGB-D Data} 

While numerous RGB-based instance segmentation models have made significant progress, instance-level segmentation with depth modality has been slower to develop. 
Earlier approaches, such as Gupta \etal~\cite{gupta2014learning} and Uhrig \etal~\cite{uhrig2016pixel}, used multi-stage and fully convolutional networks to segment instances, respectively. 
%
Later, Xu \etal~\cite{xu2020outdoor} proposed an outdoor RGB-D instance segmentation framework based on Mask R-CNN, which processed RGB and depth features separately in the encoder and then fused the two modal features to output instances.
Similarly, Xiang \etal~\cite{xiang2021learning} directly used synthetic data to learn RGB-D feature embeddings and adopted a clustering algorithm to segment invisible objects.
To advance RGB-D instance-level segmentation further, we propose a novel RGB-D SIS framework that integrates depth features into both kernel and mask branches based on the CNN-transformer architecture.

\section{DSIS Dataset}\label{dataset}

In order to create a robust RGB-D SIS benchmark and comprehensively evaluate the generalization performance of task-specific methods, we build a new exclusive RGB-D SIS dataset, namely DSIS. 

\begin{figure}
\centering
\includegraphics[width=1\linewidth]{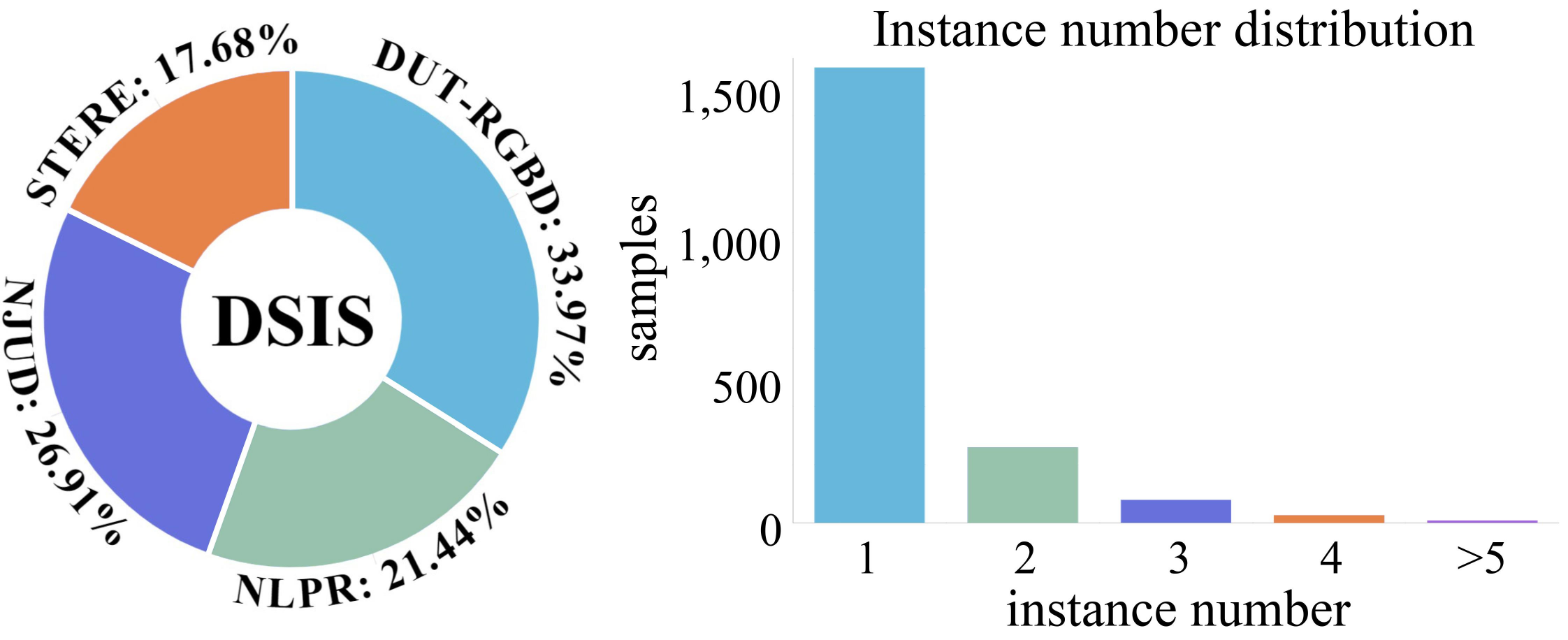}
\caption{Distribution of the DSIS dataset. Left: Distribution of image sources collected from RGB-D SOD datasets. Right: Distribution of the number of salient instances in each sample.}\label{DSIS-static}
\end{figure}

\begin{figure*}[t!]
 \centering
    \begin{overpic}[width=\linewidth]{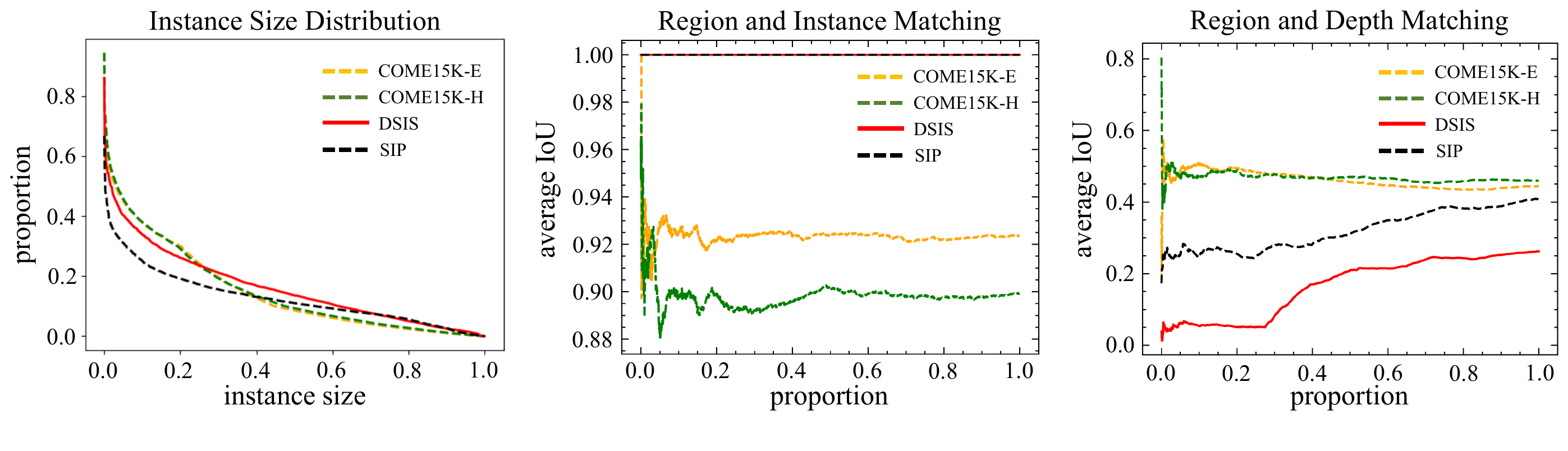}
        \put(18,0.8){\small{(a)}}
        \put(52,0.8){\small{(b)}}
        \put(85.2,0.8){\small{(c)}}
    \end{overpic}
 \caption{Comparison between the proposed DSIS and existing datasets for RGB-D SIS task. (a) Distribution of instance sizes in all test sets; (b) Comparison of the consistency between salient object-level ground truth and binarized instance-level ground truth; (c) Consistency of the salient object ground truth with the binarized depth map.}\label{dataset_analysis}
\end{figure*}

\subsection{Dataset Statistics}

To ensure a high degree of consistency in the instance-level labels and salient regions, we carefully collected about 2,500 image sources with region-level annotations and depth maps directly from the existing RGB-D SOD datasets, including NJUD~\cite{ju2015depth}, DUT-RGBD~\cite{piao2019depth}, STERE~\cite{niu2012leveraging}, and NLPR~\cite{peng2014rgbd}.
%
Subsequently, five annotators manually annotate pixel-level salient instance labels based on the existing region-level labels. 
After that, we adopt the majority voting scheme~\cite{fan2022salient} to retain consensus labels while discarding ambiguous images. 
We gather a total of 1,940 images accompanied by bounding boxes, salient objects, and salient instance annotations. 
The final distribution of data sources is provided on the left of \figref{DSIS-static}.
All images with instance-level ground truth are used for the test set to enrich the evaluation diversity.
As shown in the right of \figref{DSIS-static}, we also gathered the distribution of the number of salient instances in each image.

\begin{figure}
    \small
    \renewcommand{\tabcolsep}{6pt} 
    \renewcommand{\arraystretch}{1.2} 
    \centering
    \begin{tabular}{cc}
        \makecell[c]{\includegraphics[width=0.35\linewidth,height=0.35\linewidth]{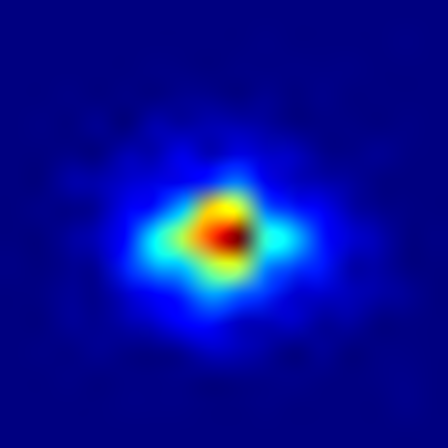}} &
        \makecell[c]{\includegraphics[width=0.35\linewidth,height=0.35\linewidth]{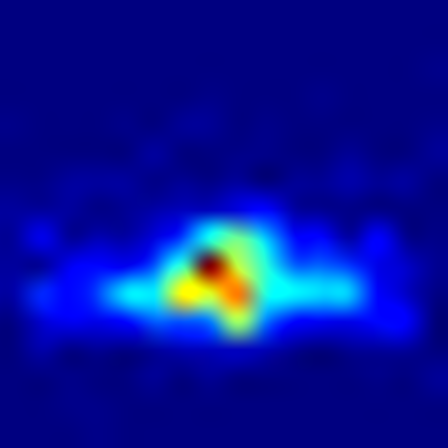}}
        \\
         \textbf{DSIS} &
         SIP 
        \\
        \makecell[c]{\includegraphics[width=0.35\linewidth,height=0.35\linewidth]{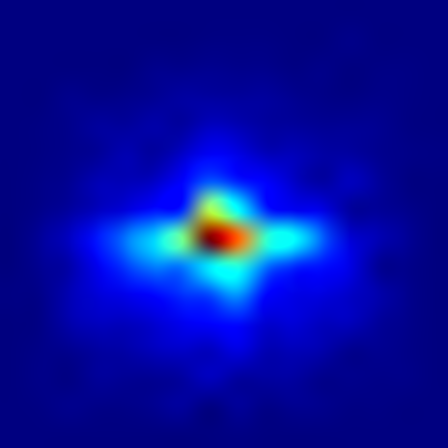}} &
        \makecell[c]{\includegraphics[width=0.35\linewidth,height=0.35\linewidth]{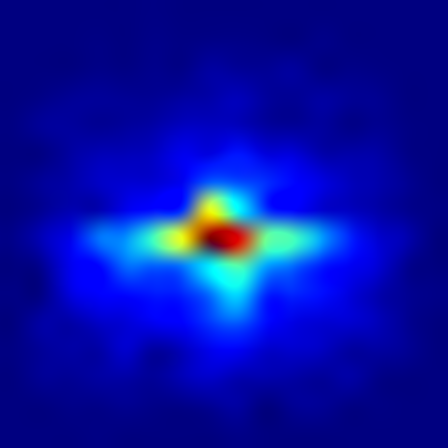}}
        \\
        COME15K-E &
        COME15K-H
        \\
    \end{tabular}
    \caption{Instance center bias of RGB-D SIS datasets.}
    \label{center-dist}
\end{figure}

Based on the existing SOD ground truth, we performed fine-grained instance-level annotations, resulting in a more comprehensive evaluation of RGB-D SIS models.
Examples of the typical annotated instances are exhibited in \figref{DSIS-show}.
We can see that our DSIS dataset contains a wide variety of scenarios and a diversity of instance appearances, including complex backgrounds, illumination changes, low contrast, high edge complexity, and so on.
In various scenarios, we have provided elaborate labels of salient instances, \eg, low-contrast airplanes, snakes under illumination changes, and vimineous bow and arrow in the column of the intricate edge.
The diversity of salient instances will bring a more comprehensive evaluation of RGB-D saliency detection models.
Compared to COME15K, the proposed DSIS guarantees consistency of region-level labels and instance-level labels, while some salient instance labels of COME15K are not in the salient regions. 
Staying ahead of SIP, our dataset has a greater variety of scenarios with multiple categories of instances. 
By releasing DSIS, we hope to facilitate further research and development in the field of RGB-D SIS.

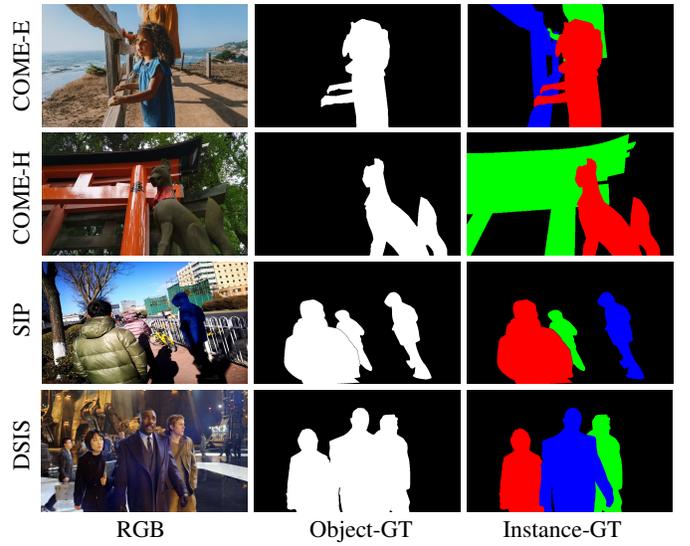
\begin{figure}[t!]
	\centering
    \begin{overpic}[width=\linewidth]{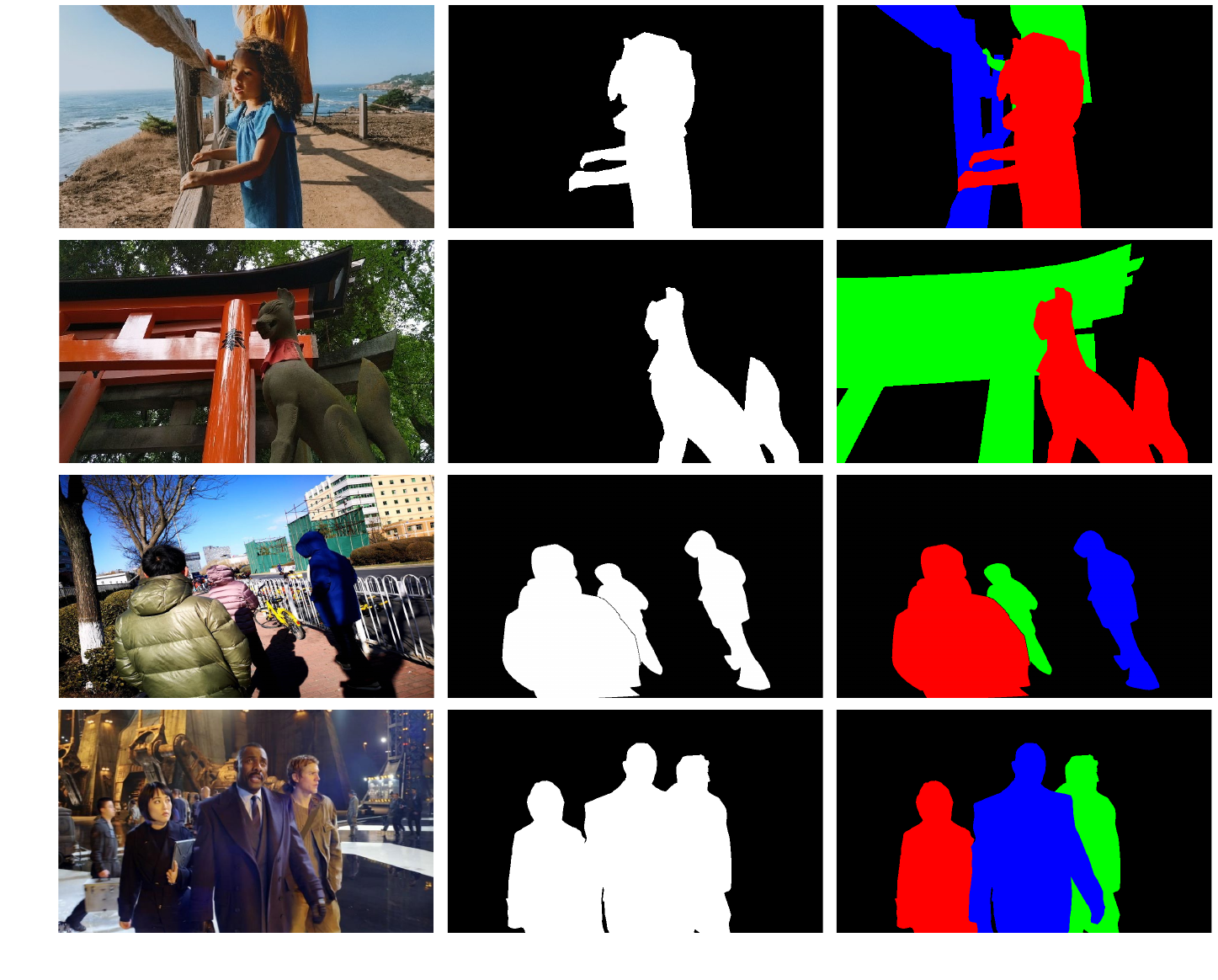}
        \put(0.5,64){\rotatebox{90}{\small{COME-E}}}
        \put(0.5,44){\rotatebox{90}{\small{COME-H}}}
        \put(0.5,30){\rotatebox{90}{\small{SIP}}}
        \put(0.5,10){\rotatebox{90}{\small{DSIS}}}
        \put(16,0){\small{RGB}}
        \put(45,0){\small{Object-GT}}
        \put(74,0){\small{Instance-GT}}
    \end{overpic}
	\caption{Comparison of the consistency between salient object-level ground truth and instance-level ground truth in RGB-D saliency datasets.}\label{inconsistency_come15k}
\end{figure}

\subsection{Dataset Analysis and Comparison}

\subsubsection{Center Bias}
The distribution of instance centers is a crucial metric that can reveal whether a dataset suffers from center bias, \ie, whether salient instances tend to be biased toward the center of the scene. 
A more dispersed distribution of instance centers indicates greater instance diversity in the dataset. 
\figref{center-dist} demonstrates that all RGB-D SIS datasets suffer from less center bias in different degrees.
Specifically, the instance centers in the COME15K-E and COME15K-H test sets \cite{zhang2021rgb} tend to be biased in the cross direction, while the instance distribution of the SIP dataset \cite{fan2020rethinking} is concentrated mainly in the lower center.
In comparison, the instance centers in our DSIS dataset are diffused more uniformly in all directions.

\begin{figure*}
\centering
\includegraphics[width=1\linewidth]{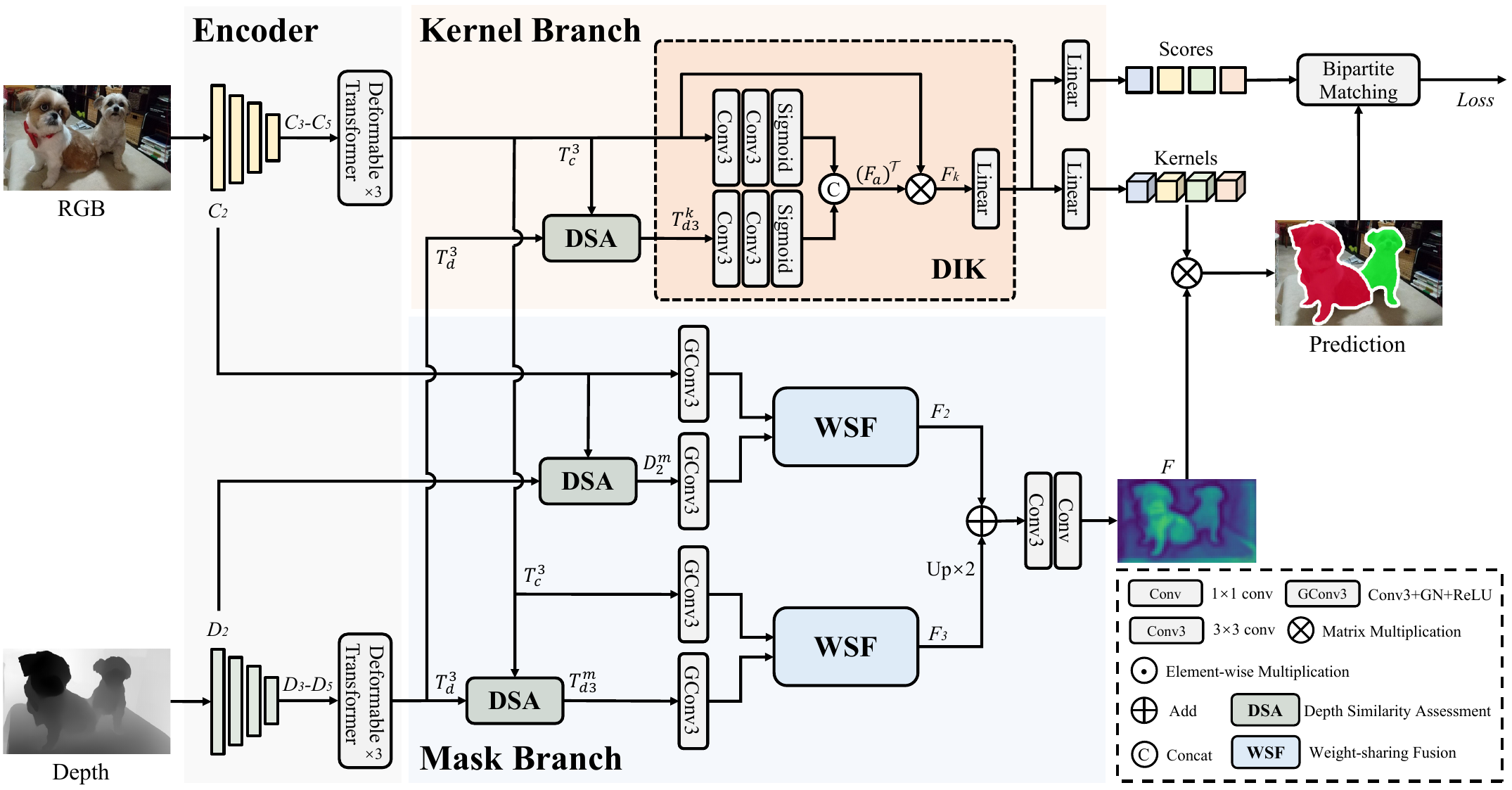}
\caption{Detailed framework of the proposed~\ourmodel~for RGB-D SIS. \ourmodel~consists of dual parallel branches: kernel and mask. 
In the kernel branch, the dynamic interactive kernel (DIK) module (\secref{kernelbranch}) is designed to interact with RGB and depth features to generate instance-aware kernels; in the mask branch, our proposed weight-sharing fusion (WSF) (\secref{maskbranch}) identifies and calibrates cross-modal information at each level and then integrates multiscale low- and high-level features to achieve the mask feature. The depth similarity assessment (DSA) (\secref{dsa}) is embedded to evaluate the similarity of RGB and depth modal so as to calibrate the depth information.
}\label{framework}
\end{figure*}

\subsubsection{Instance Size Distribution}
The distribution of instance sizes is also an important factor in evaluating the quality of a dataset since it provides insights into the variation and diversity of instances present in the images. 
The general metric for measuring instance sizes is the ratio of instance pixels to the total image pixels \cite{li2014secrets}.
As depicted in \figref{dataset_analysis} (a), we observe that the distribution of instance sizes in the DSIS and COME15K datasets cover a wide range, while the SIP dataset displays less variation in instance sizes. 
Notably, our DSIS dataset exhibits a higher proportion of medium-sized instances.
On the other hand, the COME15K dataset primarily comprises smaller-sized instances. 

\subsubsection{Objects/Instances Consistency}
As is well known, the SIS task is defined as the further partitioning of salient instances within salient regions. 
To ensure the accuracy of salient instance annotations, it is crucial to label them within the salient object-level ground truth. 
To verify the consistency between the two levels of labeling, we binarized the salient instance labels and calculated the average Intersection over Union (IoU) between the binarized instance labels and salient object labels. 
The matching curves in \figref{dataset_analysis} (b) illustrate that object- and instance-level ground truth in both SIP and DSIS are absolutely aligned, while the average IoU values in the two COME15K test sets have decreased to varying degrees. 
This implies that the salient instance labels in COME15K are not completely salient, leading to a decrease in the quality of instance labels.
\figref{inconsistency_come15k} provides typical samples that visually illustrate the inconsistency between the object- and instance-level ground truth in COME15K.

\subsubsection{Depth/Saliency Consistency}
As an RGB-D instance-level saliency detection task, the consistency between the structural information provided by the depth maps and the salient object labels can assess the difficulty of the RGB-D datasets. 
In this regard, we make a simple binarization of the depth maps for a more direct consistency comparison with the salient object labels.
Specifically, we first adopt the OTSU algorithm \cite{otsu1979threshold} to binarize the depth maps. 
Then, the foreground and background area is reversed to keep the indexes of salient and spatially closer objects consistent.
 \figref{dataset_analysis} (c) shows the matching curves between the binarized depth maps and salient object labels in all datasets. 
It is interesting to note that the matching degree between the two modalities is below 50\% in all datasets, which also illustrates that proximity to the camera does not necessarily indicate saliency.
From this perspective, it is essential to calibrate and purify the depth information in the RGB-D segmentation model.
The consistency between the depth map and saliency labels in our DSIS dataset is the lowest compared to other datasets, which poses a greater challenge for cross-modal models. 
We expect that our DSIS will encourage the development of more efficient RGB-D SIS models that can better filter and integrate cross-modal information.


\section{Proposed \ourmodel}

\subsection{Overall Architecture}

As shown in \figref{framework}, the whole framework can be divided into four parts: 
(1) two individual encoders are adopted to extract multi-level RGB features and depth features respectively; 
(2) a depth similarity assessment module is applied to bridge the inherent disparity between appearance and structure information and purify the depth cues;
(3) a kernel branch is introduced to interact with cross-modal features for producing instance-aware kernels; 
(4) a mask branch is designed to implicitly calibrate RGB and depth mutual information and fuse multiscale features to generate mask features.

\subsection{Model Encoder}

\ourmodel~extracts multi-level RGB features and depth modal features from two saliency encoders. For the RGB encoder, we first feed the input image $I$ into the ResNet-50~\cite{he2016deep} backbone to produce multiscale features denoted as \{$C_{i}|i$=$2, 3, 4, 5$\}. 
To enhance the global perception and provide higher-level features, a deformable transformer~\cite{zhu2020deformable} is employed to further achieve global features (\{$T_{c}^{i}|i$=$3, 4, 5$\}) following the CNN architecture. 
Specifically, $C_{3}$-$C_{5}$ are flattened and concatenated into a sequence that is used as the transformer input. 
To efficiently build the long-range dependencies, we adopt the vanilla deformable transformer encoder, which consists of a multi-head deformable self-attention module, layer normalization, and a feed-forward network (FFN). 
The RGB encoder stacks three deformable transformer layers sequentially and then restores the output sequence into the global features $T_{c}^{3}$-$T_{c}^{5}$. 

For the depth stream, the CNN backbone has the same structure as the RGB encoder but without sharing weights. For convenience, the output features of the depth encoder are denoted as \{$D_{i}|i$=$2, 3, 4, 5$\}. 
Three deformable transformer layers are also embedded in the depth encoder to produce high-level depth feature representations, \ie, $T_{d}^{3}$-$T_{d}^{5}$.  
The multilevel feature maps output from the two modal encoders are fed into the kernel and mask branches for further interaction.

\subsection{Kernel Branch}\label{kernelbranch}

The kernel branch aims to integrate the RGB and depth features to discover distinctive cross-modal embeddings and generate a set of instance-aware kernels. 
To this end, we introduce a DIK module that exploits two modality-specific features to facilitate cross-modal interactions toward enhancing the granularity of instance-aware points.
As illustrated in \figref{framework}, the RGB feature $T_{c}^{3}\in\mathbb{R}^{h\times w\times c}$ and the verified depth feature $T_{d3}^{k} \in\mathbb{R}^{h\times w\times c}$ after the DSA module (introduced in \secref{dsa}) are input to the DIK module.
Here, $h$, $w$, and $c$ refer to the height, width, and channel dimensions of the feature map.
In the beginning, we concatenate two-channel fixed coordinates with $T_{c}^{3}$ and $T_{d3}^{k}$ to preserve spatial locations in accordance with \cite{liu2018intriguing}. 
Bilaterally, the two modal features are operated by two successive $3\times3$ convolutional layers. 
In the latter instance-aware convolution, the dimension $c$ of the two modal features is decreased from 256 to the number of salient kernels $N$ in order to prioritize diverse potentially instance-aware embeddings. 
Then, a sigmoid function and a concatenation operation are added to generate a cross-modal instance-aware attention map $F_{a}\in\mathbb{R}^{h\times w\times N}$. This process can be formulated as
\begin{equation}
\begin{split}
F_{a}=Cat[\mathcal{S}(\mathcal{C}_{3}(\mathcal{C}_{3}(T_{c}^{3})));
\mathcal{S}(\mathcal{C}_{3}(\mathcal{C}_{3}(T_{d3}^{k})))],
\end{split}
\end{equation}
where $\mathcal{S}$ represents the sigmoid function, $\mathcal{C}_{3}$ is a 3$\times$3 convolution, and $Cat[;]$ is the concatenation operation. 
According to the guidance of the attention map, $F_{a}\in\mathbb{R}^{h\times w\times N}$ can be viewed as the key value that is used to perform matrix multiplication with the RGB feature $T_{c}^{3}\in\mathbb{R}^{h\times w\times c}$ to obtain instance-aware embeddings $F_{k}\in\mathbb{R}^{{N}\times{c}}$. 
This operation is described as
\begin{equation}
F_{k}=(F_{a})^{\mathcal{T}}\otimes T_{c}^{3},
\end{equation}
where $\otimes$ denotes matrix multiplication and $(\cdot)^{\mathcal{T}}$ is the transpose operation. 
Unlike using the cross-attention operation~\cite{carion2020end,cheng2022masked} to obtain kernels, our DIK module is more lightweight that boasts adaptability to diverse multi-modal input settings.
%
Finally, given $F_{k}$, we can obtain a set of confidence scores $S^{{N}\times{1}}$ and instance-aware kernels $K^{{N}\times{c}}$ by a shared linear layer and two separate linear layers.

\begin{figure}
\centering
\includegraphics[width=1\linewidth]{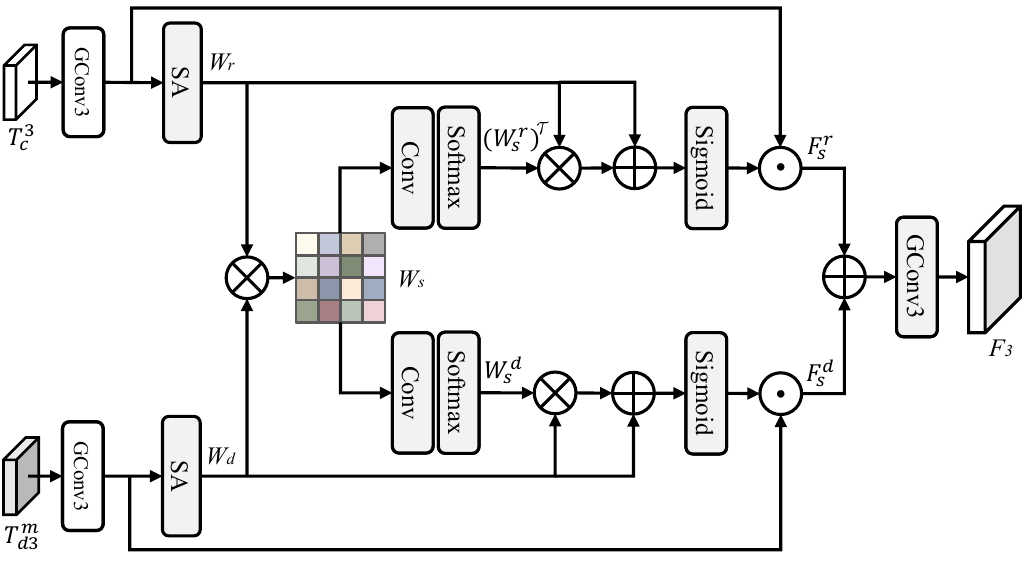}
\caption{Structure of the proposed WSF module. The same operation is performed for the other scale ($C_{2}, D_{2}^{m}$). SA is the spatial attention (SA)~\cite{woo2018cbam} to generate spatial weights.
}\label{WSF}
\end{figure}

\subsection{Mask Branch}\label{maskbranch}

In tandem with the kernel branch, the mask branch is brought out to discover and calibrate distinctive information from cross-modal cues and fuse multiscale low- and high-level features, ultimately yielding a shared mask feature. 
For the input of the mask branch, we employ the low-level feature $C_{2}$ from the CNN backbone and the high-level feature $T_{c}^{3}$ from the deformable transformer in the RGB encoder to provide a rich appearance feature representation. 
In concurrence with $C_{2}$ and $T_{c}^{3}$, depth features $D_{2}$ and $T_{d}^{3}$ from the depth encoder are also applied to enhance structural information and maintain the internal consistency of salient regions. 
We divide different modality features of the same scale into the same group (\ie, $\{C_{2}, D_{2}\}, \{T_{c}^{3}, T_{d}^{3}\}$). 
The details of the mask branch are described at the bottom of \figref{framework}.  
Similar to the kernel branch, the input depth features $D_{2}$ and $T_{d}^{3}$ are also passed through the DSA module to obtain purified depth features $D_{2}^{m}$ and $T_{d3}^{m}$.
After a sequential operation (3$\times$3 convolution, group normalization, and a ReLU function), the RGB and depth features are fed into the proposed WSF, which is designed to efficiently exploit complementary cross-modal features at different levels for a shared representation. 

As illustrated in \figref{WSF}, the WSF module aims to implicitly calibrate and integrate cross-modal features based on a shared affinity weight which establishes strong correlations and reveals mutual information between the two modalities.
%
Toward this goal, the two modal features first pass through spatial attention (SA) \cite{woo2018cbam} to produce individual spatial attention weights $W_{r}\in\mathbb{R}^{ h \times w}$ and $W_{d}\in\mathbb{R}^{h \times w}$.
Given the depth feature $T_{d3}^{m}$, the process of SA can be described as
\begin{equation}
W_{d}=\mathcal{S}(\mathcal{C}_{7}(Cat[\mathcal{A}(T_{d3}^{m});\mathcal{M}(T_{d3}^{m})]),
\end{equation}
where $\mathcal{A}$ denotes the average-pooling operation and $\mathcal{M}$ is the max-pooling operation.
Consistently, $D_{2}^{m}$, $C_{2}$, and $T_{c}^{3}$ pass through the same operation to obtain corresponding spatial attention weights.
For better aggregation and calibration of cross-modal features, the shared affinity weight $W_{s}\in\mathbb{R}^{h \times h}$ is generated from the spatial weight maps $W_{r}$ and $W_{d}$, which is calculated as 
\begin{equation}
W_{s} = W_{r}\otimes (W_{d})^{\mathcal{T}}.
\end{equation}
It is worth noting that the shared affinity matrix $W_{s}$ has a small size of $h\times h$, which avoids computing the pixel-to-pixel affinity weights ($hw\times hw$)~\cite{wang2018non, cong2022cir, liu2021learning}.
Moreover, the affinity weight with a smaller size can greatly decrease the computational cost of subsequent operations in WSF (see \secref{ablation}).
Once receiving the affinity weight $W_{s}$, we can generate the modality-specific weights $W_{s}^{r}\in\mathbb{R}^{h \times h}$ and $ W_{s}^{d}\in\mathbb{R}^{h \times h}$ 
using two 1$\times$1 convolution layers and the softmax function.
Afterward, $W_{s}^{r}$ and $W_{s}^{d}$ are integrated with the previous spatial weights $W_{r}$ and $W_{d}$ by the matrix multiplication with the residual operation to enhance the modality-specific spatial attention and preserve the original modal features. 
An element-wise multiplication is followed to achieve the re-weighted RGB feature $F_{s}^{r}$ and depth feature $F_{s}^{d}$.
For instance, given the input features $T_{c}^{3}$ and $T_{d3}^{m}$, the two modal re-weighted features can be achieved as
\begin{equation}
\begin{cases}
F_{s}^{r}=\mathcal{G}_{3}(T_{c}^{3})\odot \mathcal{S}(W_{r}+(W_{s}^{r})^{\mathcal{T}}\otimes W_{r}) \\ 
F_{s}^{d}=\mathcal{G}_{3}(T_{d3}^{m})\odot \mathcal{S}(W_{d}+W_{s}^{d}\otimes W_{d})
\end{cases}
\end{equation}
where $\odot$ is element-wise multiplication, $\mathcal{G}_{3}$ denotes a sequential operation that contains a 3$\times$3 convolution followed by group normalization, and a ReLU function. 
In this way, we can add re-weighted RGB and depth features ($F_{s}^{r}$ and $F_{s}^{d}$) to obtain the fused cross-modal feature map $F_{3}$ with a scale of 1/8. Similarly, we can also take $C_{2}$ and $D_{2}^{m}$ as inputs and use WSF to obtain a finer fused feature $F_{2}$ with a scale of 1/4.

Integrating the fused features across different scales is essential to generate a shared mask feature. 
In this regard, we concisely upsample $F_{3}$ to the 1/4 scale and add it to $F_{2}$. 
A 3$\times$3 convolution followed by a 1$\times$1 convolution is applied to produce the unified mask feature $F$. 
Finally, the instance-aware kernels $K\in\mathbb{R}^{{N}\times{c}}$ and the mask feature $F\in\mathbb{R}^{{c}\times {h}\times {w}}$ are fed into the dynamic convolution operation to output salient instance masks $M\in\mathbb{R}^{{N}\times{h}\times{w}} = K \otimes F$.

\begin{figure}
\centering
\includegraphics[width=1\linewidth]{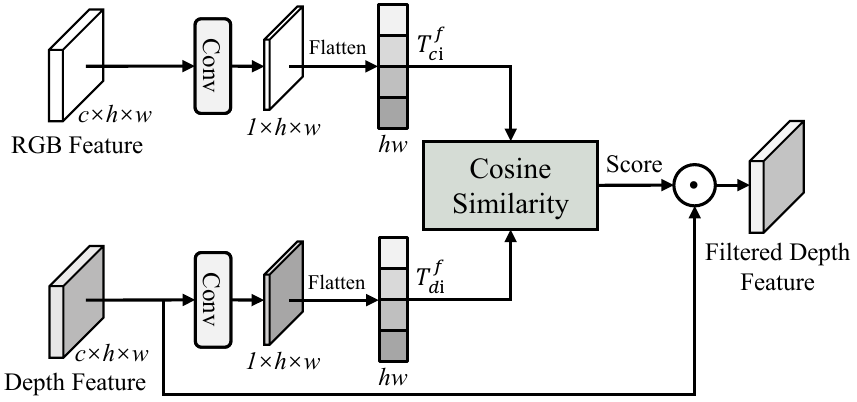}
\caption{Diagram of the proposed DSA module.}\label{cosine_sim}
\end{figure}

\subsection{DSA Module}\label{dsa}

As is well known, structural information and salient cues are not strongly correlated. Low-quality depth maps contain noisy information that interferes with the prediction of salient instances. 
For this purpose, we propose a DSA module that develops a similarity score by evaluating the correlation between RGB and depth information to adaptively weigh the depth features. 
In~\ourmodel, the DSA module is embedded before the kernel branch and the mask branch to calibrate the depth features. The diagram of our DSA module is shown in \figref{cosine_sim}.
Symmetrically, two modal features from the encoder pass through a 1$\times$1 convolution to generate an attention map, respectively. 
Then, the RGB and depth attention maps are flattened to calculate their attention similarity. 
Notably, we adopt the cosine distance to calculate the similarity score due to its high sensitivity to directional and spatial considerations. 
Giving the flattened attention maps $T_{ci}^{f}$ and $T_{di}^{f}$, the similarity score $V_{s}$ is computed as
\begin{equation}
V_{s}=1-(1-\frac{T_{di}^{f}\odot T_{ci}^{f}}{\left \| T_{di}^{f} \right \| \left \| T_{ci}^{f} \right \|})/2.
\end{equation}
The range of the similarity score $V_{s}$ is [0, 1]. The higher the value of $V_{s}$, the stronger the consistency of different modal features.
Finally, we multiply the similarity score with the input depth features to optimize the fusion proportion of depth information.

\begin{table*}[!t]
\begin{center}
\caption{Quantitative comparisons with 19 task-related methods for RGB-D SIS.}
\label{tab:SOTA}
\footnotesize
    \renewcommand{\arraystretch}{1.2}
    \renewcommand{\tabcolsep}{1.95mm}
\begin{tabular}{c|l|l|ccc|ccc|ccc|ccc}
\toprule
\multirow{2}{*}{Task} &\multirow{2}{*}{Method}  &\multirow{2}{*}{Volume}  &\multicolumn{3}{c|}{COME15K-E~\cite{zhang2021rgb}}    & \multicolumn{3}{c|}{COME15K-H~\cite{zhang2021rgb}} & \multicolumn{3}{c|}{DSIS} & \multicolumn{3}{c}{SIP~\cite{fan2020rethinking}} \\ 
\cline{4-15} 
          &            &        &       AP     &      AP$_{50}$      &     AP$_{70}$    &   AP     &  AP$_{50}$    & AP$_{70}$    & AP   & AP$_{50}$ & AP$_{70}$   & AP   & AP$_{50}$ & AP$_{70}$  \\ 
\hline\hline
\multirow{4}{*}{SIS}
&S4Net \cite{fan2019s4net}   &  CVPR19     &  43.7   &  68.0   &   52.5    &   37.1     &       60.9    &            43.2     &     58.3   &   81.9   &      71.8   &   49.6   &   76.0  &  63.7       \\
&RDPNet  \cite{wu2021regularized}  & TIP21      &  49.8   & 72.2   &  59.5     &  42.1      &     65.2      &      49.7    &    66.1     &  87.2      &  80.1          &     59.0     &   80.1    &     74.1       \\
&OQTR \cite{pei2022transformer}  & TMM22    &   48.7  &  71.1   &  58.3    &   42.7   &   65.9   &    50.5   &   63.1   &   85.9   &   77.0    &  59.9  &    83.1   &  76.3   \\ \hline
\multirow{1}{*}{CIS}
&OSFormer \cite{pei2022osformer}    & ECCV22     & 53.0  &  71.9     &   61.3    &   45.8   &   66.3  &     52.4  &  67.9  &   86.3   &  78.3  & 63.2 & 80.5  &   74.6  \\ \hline
\multirow{16}{*}{GIS}
&Mask R-CNN \cite{he2017mask}  &   ICCV17       &  48.8   &  71.2     &   58.6     &   42.2      &     65.7      &  50.8      &  65.6      &  86.8    &    80.2     &   57.9    &  79.8       &   73.3    \\
&MS R-CNN \cite{huang2019mask}  &  CVPR19    &  49.7    &  70.4     &   58.9     &   42.3      &     63.8      &      50.0     &   66.8    &      87.3   &    80.3     &   60.0    &   79.8    &   73.3    \\
&Cascade R-CNN \cite{cai2019cascade}  & TPAMI19     &  49.4   &   69.8    &    58.7    &     41.9    &      64.3    &      49.8     &   66.4    &    87.3    &    79.9    &   59.1    &   79.1    &   73.6    \\
&CenterMask \cite{lee2020centermask}  &  CVPR20     & 49.6   &   71.6   &   59.0     &       42.5   &        65.2  &   51.0      &    65.7    &           87.6    &     79.7     &    57.6    &    79.8        &  72.3     \\
&HTC \cite{chen2019hybrid}    &  CVPR19   & 51.4  &  73.7     &   61.0     &       44.1         &     68.0       &      52.1   &    67.5    &           87.6   &     81.1     &    60.0   &     81.4       &   74.9    \\
&YOLACT \cite{bolya2019yolact}   &   ICCV19     &   48.1  &   70.7    &    56.2    &   41.4     &         66.0      &      48.5     &   62.4     &    84.1    &     73.3     &      62.0      &    82.3   &   74.7   \\
&BlendMask \cite{chen2020blendmask}    &   CVPR20    & 48.1  &   70.6    &  56.9  &     41.0    &    64.8         &   48.5    &  65.5   &     86.9   &  78.0   &   55.5    &    77.3    &  69.7  \\
&CondInst \cite{tian2020conditional}  &  ECCV20     & 49.6  &  72.0     &  59.5      &    42.8   &       66.4        &     51.0     &   65.1     &    86.8   &    79.1     &   58.6    &     79.4       &  73.3   \\
&SOLOv2 \cite{wang2021solo}   & TPAMI21     & 51.1  &  71.6  &  59.9   &    45.1    &    66.3    &   52.8       &   67.4   &    87.0   &  80.4   &  63.4    &   80.7    &  74.8   \\
&QueryInst \cite{fang2021instances}  &  CVPR21    &  51.5  &   73.1    &   61.1     &   43.9     &  67.5      &    51.8 &   67.9  &    87.5    &    80.6   &   61.3    &   81.0     &  75.1  \\
&SOTR \cite{guo2021sotr}      &   ICCV21     &  50.7   &   70.4    &       59.0   &       43.5     &       65.1      &     50.3        &     68.2     &    86.9  &  79.3  &   61.6    &   78.4     &   73.0    \\
&Mask Transfiner \cite{ke2022mask}  &   CVPR22    & 48.7  &   71.0    &   56.3     &  40.7        &      64.5     &     47.2     &   67.5     &    87.6   &    80.4     &   57.8    &    78.9    &   70.3    \\
&SparseInst \cite{cheng2022sparse}   & CVPR22    & 51.3  &  71.9   &  58.9   &    43.1  &   65.1     &    48.5  &  65.0   &   85.0  &  75.7   &  62.8  &  81.3   & 75.3  \\
&Mask2Former \cite{cheng2022masked}   &  CVPR22      &  51.5  &  67.3  &   57.6    &    44.1     &   62.4     &     49.4    &   68.3   &    85.1      &    76.7     &   66.6    &   79.3   &  75.1 \\ \hline
\multirow{1}{*}{RGB-D GIS}
& RRL [9]  &  TIP20  &   52.9    &    71.8   &     61.8   &    
45.6  &         66.1          &      53.6          &    66.1   &   87.1    &    79.6   &  61.9     &          79.1 &     74.6     \\ \hline
\rowcolor[RGB]{235,235,235}
\multirow{1}{*}{DSIS}
&\textbf{\ourmodel~(Ours)}   &  -    &   \textbf{58.0}     &    \textbf{75.8}   &     \textbf{65.6}   &                \textbf{50.7}  &         \textbf{70.4}            &        \textbf{57.3}          &    \textbf{69.3}    &    \textbf{87.8}    &    \textbf{81.6}   &  \textbf{72.1}     &          \textbf{86.6}  &     \textbf{82.9}      \\ \bottomrule
\end{tabular}
\end{center}
\end{table*}

\subsection{Loss Function}

For end-to-end training, the proposed~\ourmodel~employs bipartite matching to assign confidence scores and instance masks. 
In line with DETR~\cite{carion2020end}, we apply the Hungarian algorithm to optimize the paired matching between ground truths and predictions. 

The training loss function is defined as follows:
\begin{equation}
\mathcal{L}=\lambda_{c}\mathcal{L}_{c}+\lambda_{mask}\mathcal{L}_{mask}+\lambda_{obj}\mathcal{L}_{obj}+\lambda_{bin}\mathcal{L}_{bin},
\end{equation}
where $\mathcal{L}_{c}$ is the saliency localization loss that leverages the confidence score to determine salient instances. 
Specifically, focal loss~\cite{lin2017focal} is used in $\mathcal{L}_{c}$. 
$\mathcal{L}_{mask}$ is the segmentation mask loss combining Dice loss~\cite{milletari2016v} and pixel-wise binary cross-entropy loss. 
Inspired by~\cite{cheng2022sparse}, we also embed the binary cross-entropy loss $\mathcal{L}_{obj}$ for the IoU-aware objectness. 
In addition, we employ the auxiliary loss $\mathcal{L}_{bin}$ to enhance the supervision of salient regions, which can be formulated as
\begin{equation}
\begin{split}
\mathcal{L}_{bin}  =\sum\limits_{r\in\left\{C_{2}, T_{3}\right\}}\lambda_{r}\mathcal{L}_{bce}(\mathcal{P}_{r},\mathcal{R}_{o}) + \\
\sum\limits_{d\in\left\{D_{2}, D_{3}\right\}}\lambda_{d}\mathcal{L}_{bce}(\mathcal{P}_{d},\mathcal{R}_{o}),
\end{split}
\end{equation}
where $\mathcal{L}_{bce}$ is the binary cross-entropy loss, $\mathcal{R}_{o}$ denotes the region-level ground truth derived from the instance-level label. 
$\mathcal{P}_{r}$ and $\mathcal{P}_{d}$ denote the region-level predictions that are generated from the RGB and depth feature maps after the encoder by adding a 1$\times$1 convolution followed by a sigmoid function. 
The corresponding coefficients $\lambda_{r}$ and $\lambda_{d}$ are set to 0.6 and 0.4 respectively. 
Besides, $\lambda_{c}$, $\lambda_{mask}$, $\lambda_{obj}$, and $\lambda_{bin}$ are empirically set to 2, 1, 1, and 1 respectively to balance the total loss function.


\section{Experiments}

\subsection{Datasets}

As a novel challenge, the available labels can be obtained from the instance-level annotations of RGB-D SOD datasets, including COME15K~\cite{zhang2021rgb} and SIP~\cite{fan2020rethinking}. 
To be specific, the COME15K dataset consists of 8,025 images with instance-level labels for training. 
For testing, it includes two test sets of different difficulty levels: COME15K-E with 4,600 normal samples and COME15K-H with 3,000 difficult samples. Before, Fan \etal\cite{fan2020rethinking} released an RGB-D SOD dataset with instance-level annotations, termed SIP, which embraces 929 high-resolution images for testing. 
To enrich and thoroughly evaluate the performance of RGB-D SIS methods,  we released a new RGB-D SIS dataset, called DSIS. 
DSIS consists of 1,940 samples with high-quality instance-level labels and corresponding depth maps for testing. 
Further details of DSIS are available in \secref{dataset}.
In this paper, we use the instance-level training set of COME15K to train our~\ourmodel~and evaluate the prediction results on the test sets of all datasets. 

\subsection{Evaluation Metrics} 

In accordance with previous SIS methods~\cite{wu2021regularized, pei2022transformer}, we also adopt the COCO-like evaluation metrics to evaluate RGB-D SIS models. 
The AP (average precision) score reflects the comprehensive performance of the tested method. 
AP$_{50}$ and AP$_{70}$ evaluate the predictions of salient instances more meticulously under specific mask IoU thresholds. 

\subsection{Implementation Details}

We implement~\ourmodel~based on Detectron2~\cite{wu2019detectron2} and train and test it on an RTX 3090 GPU with a batch size of 4. 
In the training stage, all models are trained for 50 epochs using the AdamW optimizer~\cite{loshchilov2017decoupled} with an initial learning rate of 2.5e-5 and a weight decay of 0.0001. The learning rate is divided by 10 at the 35\emph{th} and 45\emph{th} epochs respectively. 
Unless specially mentioned, we use the ResNet-50~\cite{he2016deep} with the pre-trained weights on ImageNet~\cite{deng2009imagenet} as the CNN backbone. All other layers are randomly initialized. 
The input RGB and depth images are resized as the length of the shorter side is 320 and the longer side is at most 480 according to the original aspect ratio.
For data augmentation, we adopt random flipping and scale jitter. 
Note that, the default number of kernels $N$ is set to 50, and the number of transformer layers in both RGB and depth encoders is 3.

\subsection{Comparison with State-of-the-art Models}

To the best of our knowledge, \ourmodel~is the first model for RGB-D SIS. 
For a comprehensive comparison, we bring 19 well-known task-related models: S4Net~\cite{fan2019s4net}, RDPNet~\cite{wu2021regularized}, and OQTR~\cite{pei2022transformer} from SIS; 
OSFormer~\cite{pei2022osformer} from camouflaged instance segmentation (CIS); 
Mask R-CNN~\cite{he2017mask}, MS R-CNN~\cite{huang2019mask}, Cascade R-CNN~\cite{cai2019cascade}, CenterMask~\cite{lee2020centermask}, HTC~\cite{chen2019hybrid}, YOLACT~\cite{bolya2019yolact}, BlendMask~\cite{chen2020blendmask}, CondInst~\cite{tian2020conditional}, SOLOv2~\cite{wang2021solo}, QueryInst~\cite{fang2021instances}, SOTR~\cite{guo2021sotr}, Mask Transfiner~\cite{ke2022mask}, SparseInst~\cite{cheng2022sparse}, and Mask2Former~\cite{cheng2022masked} from generic instance segmentation (GIS); RRL~\cite{xu2020outdoor} from RGB-D GIS.
For a further in-depth comparison, we select the top three above-mentioned models and enhance them with fused depth modality to directly compare with our model on the RGB-D SIS task. 
Notably, we uniformly use the official code to train all models on the COME15K training set and evaluate them on the test sets of COME15K, SIP, and our DSIS. 
Additionally, all models are based on the ResNet-50~\cite{he2016deep} backbone with the ImageNet~\cite{deng2009imagenet} pre-trained weights. 

\begin{table}[!t]
\begin{center}
\caption{Performance comparison with~\sArt~instance-level models after fusing depth modality.}
\label{tab:D-SOTA}
\footnotesize
    \renewcommand{\arraystretch}{1.2}
    \renewcommand{\tabcolsep}{1.1mm}
\begin{tabular}{l|ccc|ccc|c|c}
\toprule
\multirow{2}{*}{Method} &\multicolumn{3}{c|}{COME15K-E~\cite{zhang2021rgb}}    & \multicolumn{3}{c|}{COME15K-H~\cite{zhang2021rgb}} & \multirow{2}{*}{Params} & \multirow{2}{*}{FPS} \\ 
\cline{2-7} 
     &       AP     &      AP$_{50}$      &     AP$_{70}$    &   AP     &  AP$_{50}$    & AP$_{70}$    &    &    \\ 
\hline\hline
OSFormer$^{*}$     & 50.9  & 71.0    &  58.6   &   43.3   &  65.2  & 50.2  &  78.9M &  28.1   \\ 
Mask2Former$^{*}$    &  55.9 &   72.7  &   62.4  &   48.3   &  66.9  &   54.1   & \textbf{68.3}M  &  28.0  \\
QueryInst$^{*}$   &  51.5 &   73.4  &   61.4  &   44.6   &  67.7  &   52.8   &  199.8M &  16.9 \\
\rowcolor[RGB]{235,235,235}
\textbf{\ourmodel~(Ours)}   &  \textbf{58.0}     &    \textbf{75.8}   &     \textbf{65.6}   &                \textbf{50.7}  &         \textbf{70.4}            &        \textbf{57.3}      &  78.9M   &   \textbf{35.9}   \\ \bottomrule
\end{tabular}
\end{center}
\end{table}

\begin{figure*}[t]
    \small
    \renewcommand{\tabcolsep}{1.7pt} 
    \renewcommand{\arraystretch}{1} 
    \centering
    \begin{tabular}{cccccc}
        \makecell[c]{\includegraphics[width=0.16\linewidth,height=0.098\linewidth]{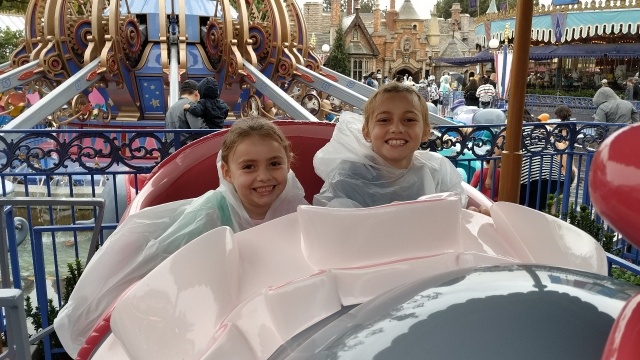}} &
        \makecell[c]{\includegraphics[width=0.16\linewidth,height=0.098\linewidth]{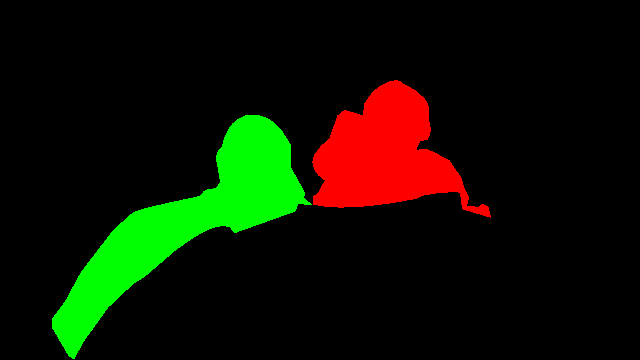}} &
        \makecell[c]{\includegraphics[width=0.16\linewidth,height=0.098\linewidth]{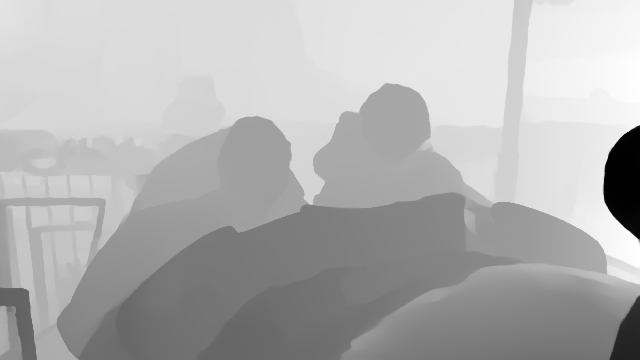}} &
        \makecell[c]{\includegraphics[width=0.16\linewidth,height=0.098\linewidth]{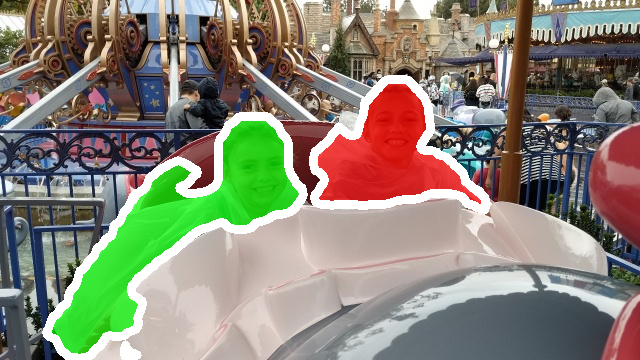}} &
        \makecell[c]{\includegraphics[width=0.16\linewidth,height=0.098\linewidth]{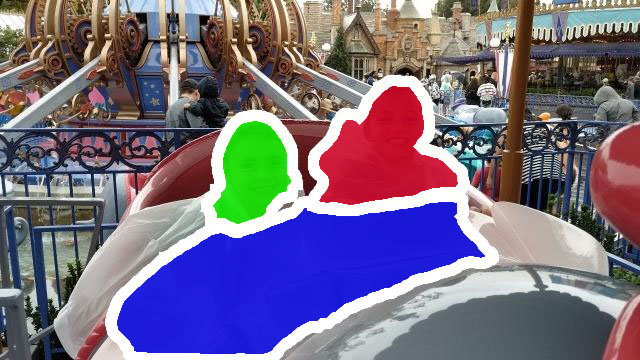}} &
        \makecell[c]{\includegraphics[width=0.16\linewidth,height=0.098\linewidth]{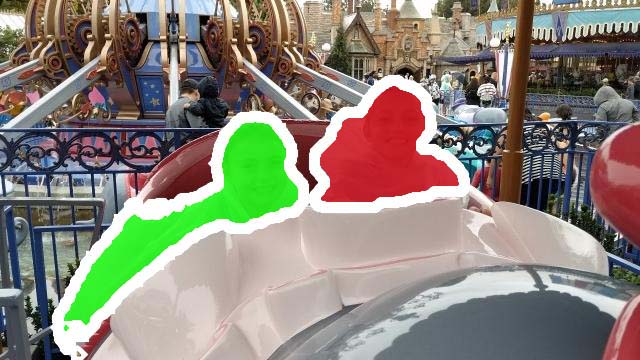}}
        \\
        \makecell[c]{\includegraphics[width=0.16\linewidth,height=0.098\linewidth]{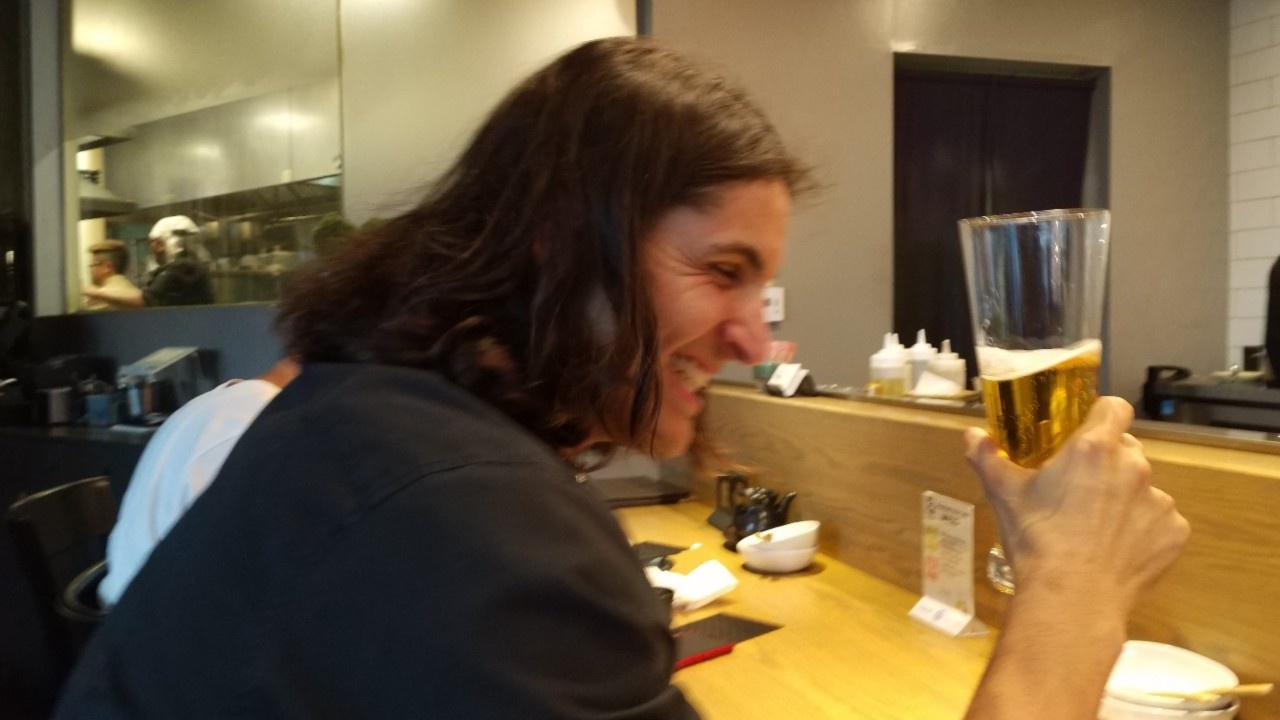}} &
        \makecell[c]{\includegraphics[width=0.16\linewidth,height=0.098\linewidth]{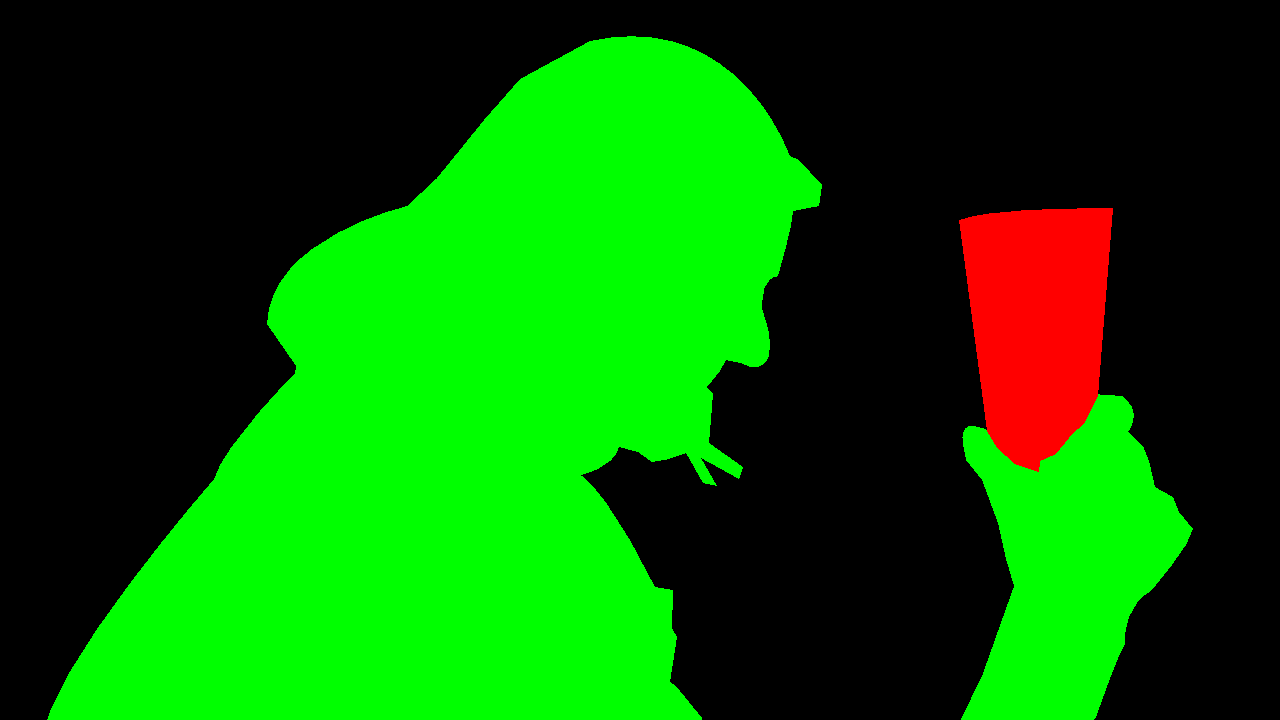}} &
        \makecell[c]{\includegraphics[width=0.16\linewidth,height=0.098\linewidth]{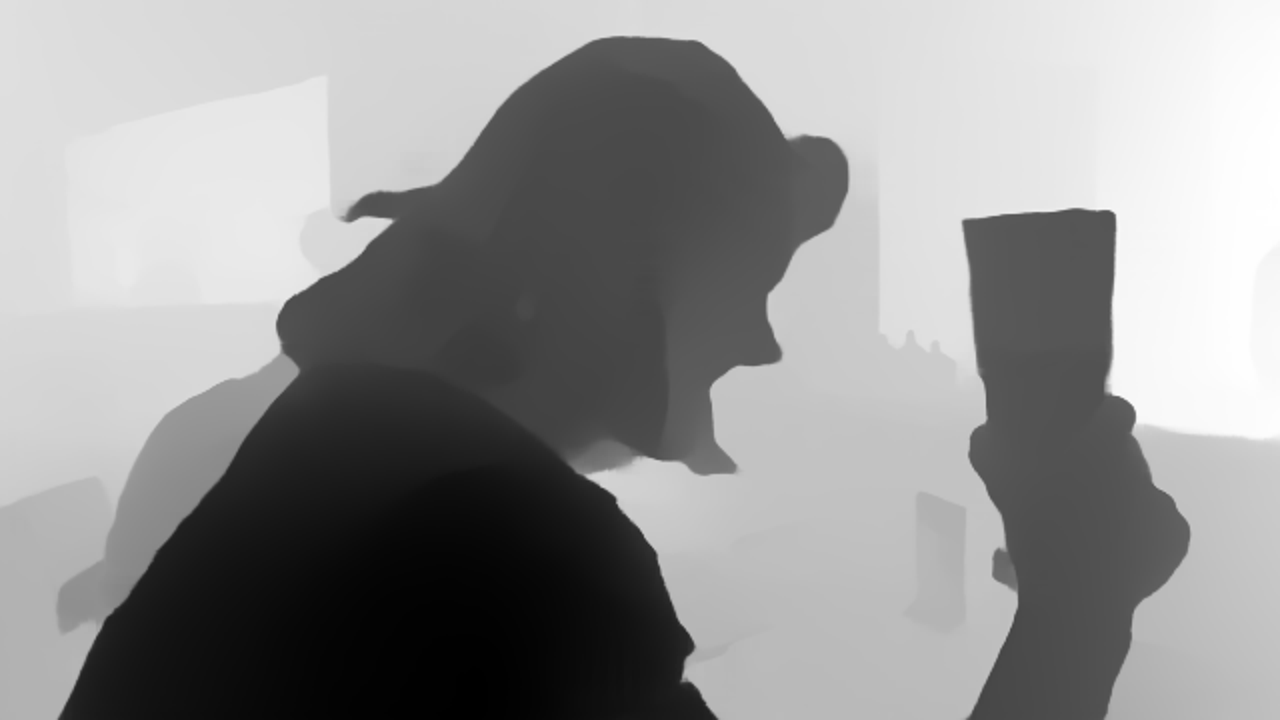}} &
        \makecell[c]{\includegraphics[width=0.16\linewidth,height=0.098\linewidth]{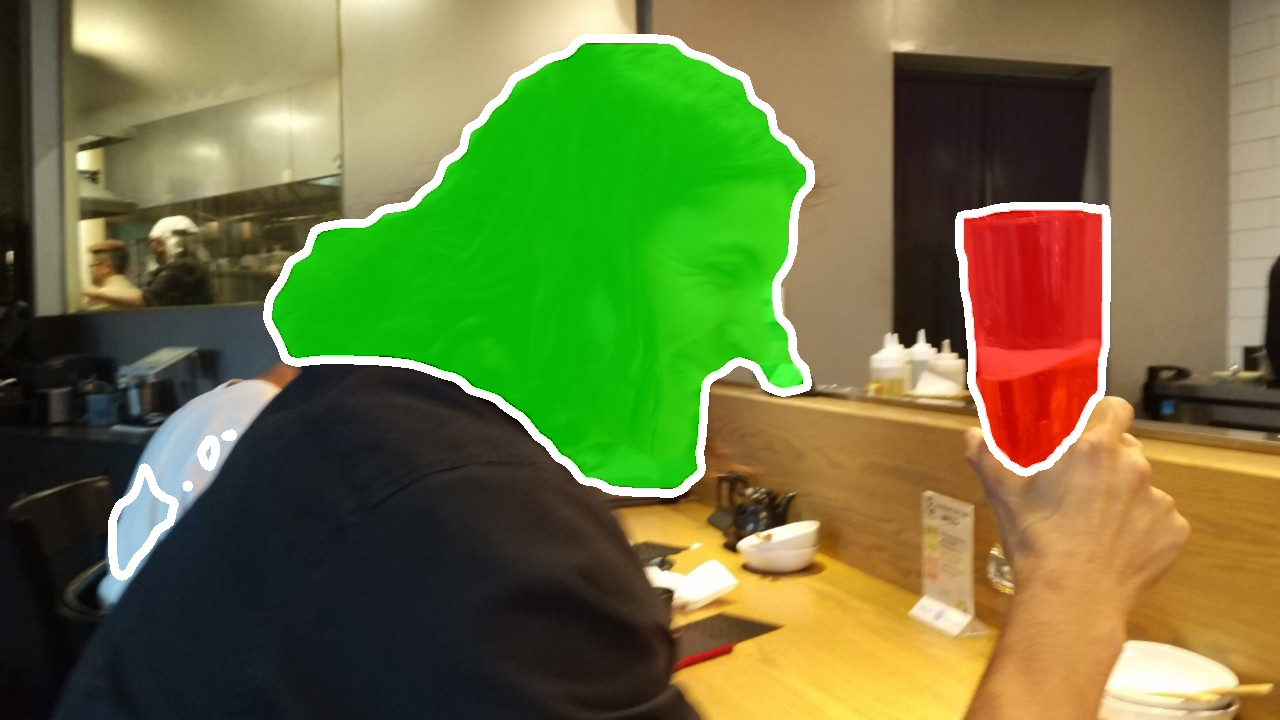}} &
        \makecell[c]{\includegraphics[width=0.16\linewidth,height=0.098\linewidth]{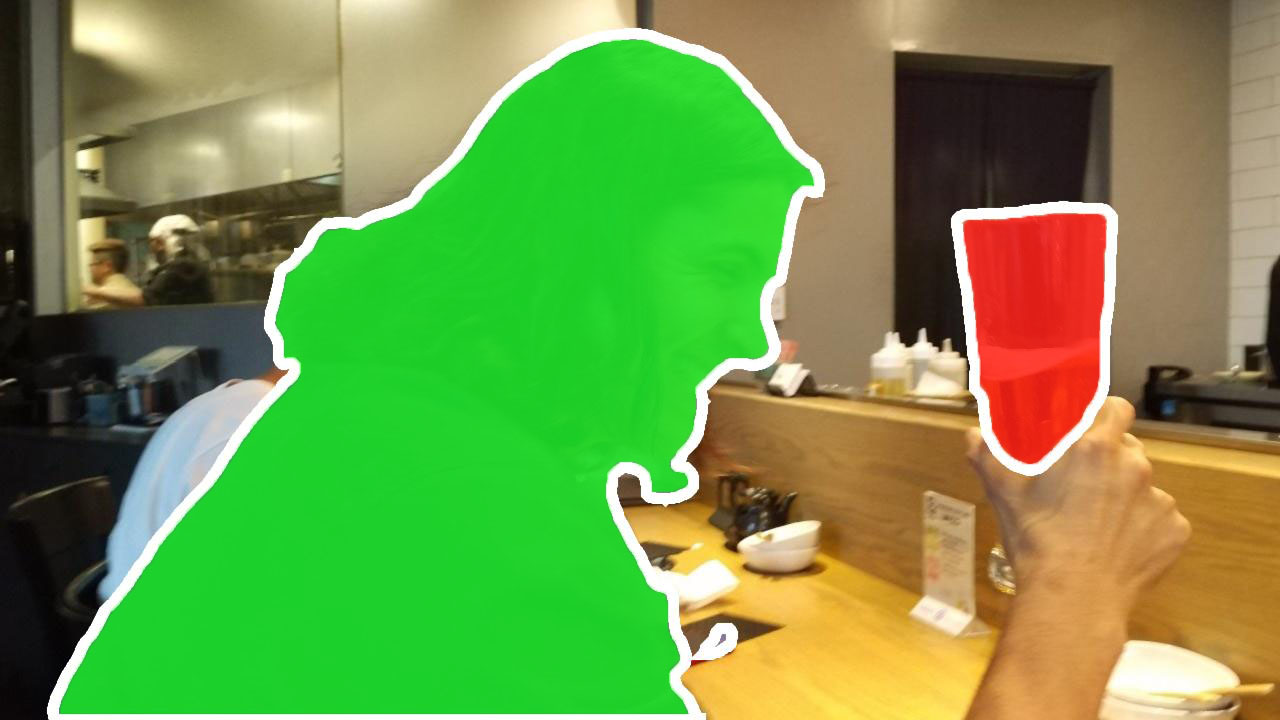}} &
        \makecell[c]{\includegraphics[width=0.16\linewidth,height=0.098\linewidth]{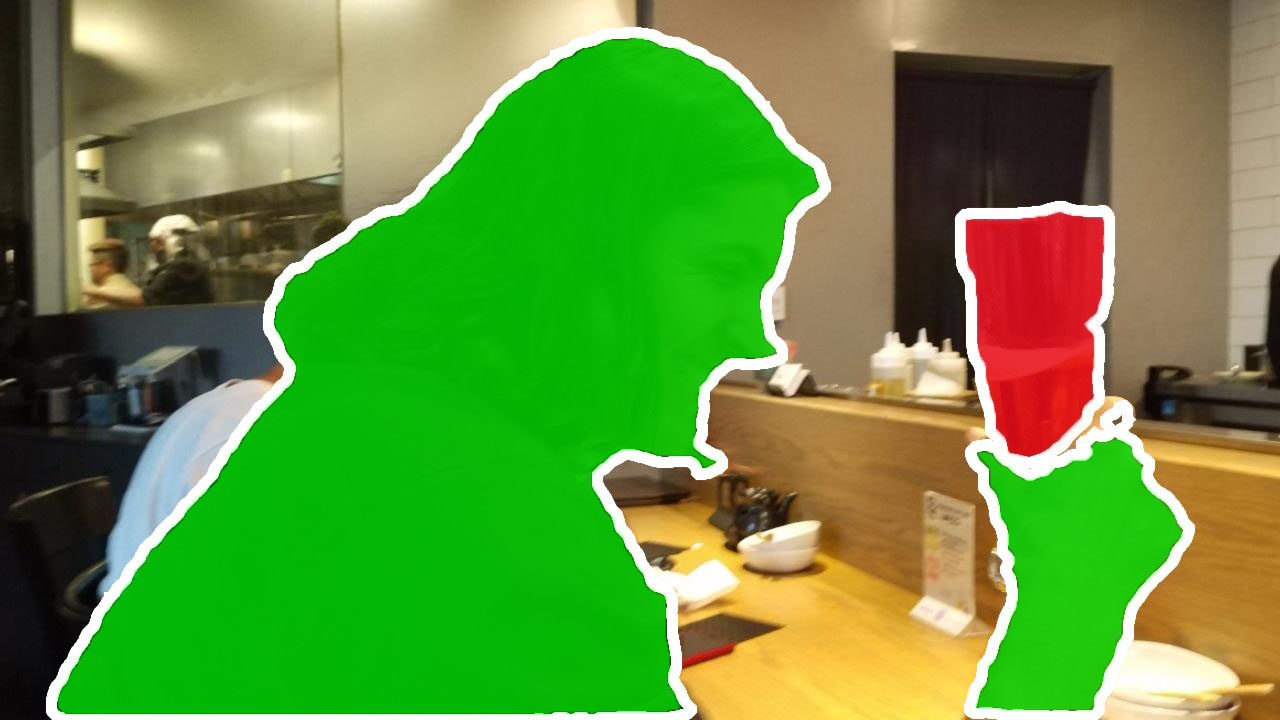}}
        \\
        \makecell[c]{\includegraphics[width=0.16\linewidth,height=0.098\linewidth]{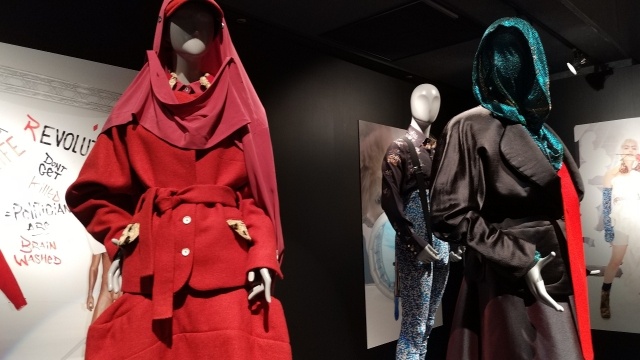}} &
        \makecell[c]{\includegraphics[width=0.16\linewidth,height=0.098\linewidth]{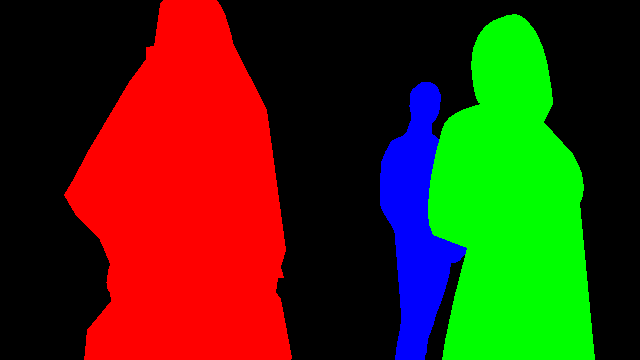}} &
        \makecell[c]{\includegraphics[width=0.16\linewidth,height=0.098\linewidth]{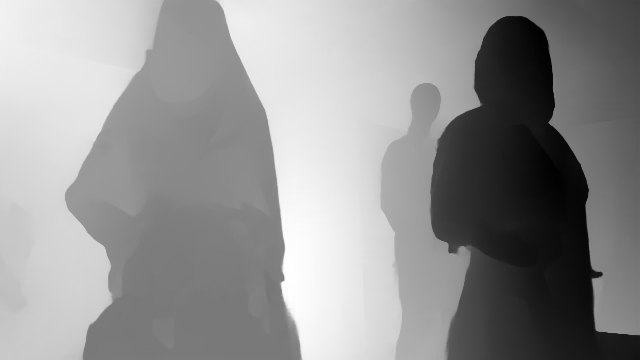}} &
        \makecell[c]{\includegraphics[width=0.16\linewidth,height=0.098\linewidth]{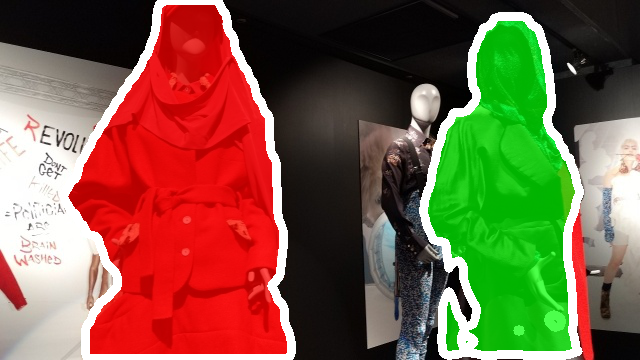}} &
        \makecell[c]{\includegraphics[width=0.16\linewidth,height=0.098\linewidth]{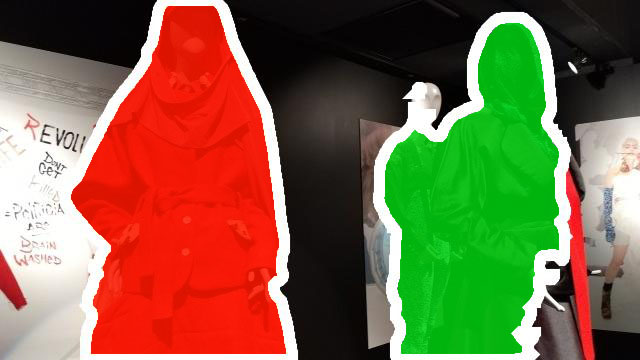}} &
        \makecell[c]{\includegraphics[width=0.16\linewidth,height=0.098\linewidth]{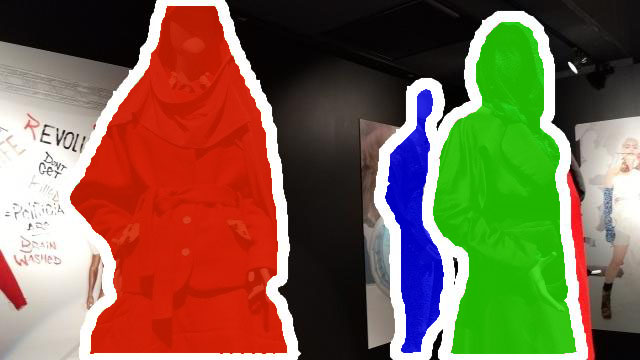}}
        \\
        \makecell[c]{\includegraphics[width=0.16\linewidth,height=0.098\linewidth]{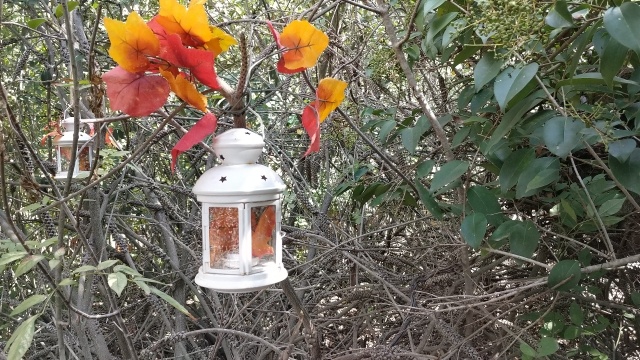}} &
        \makecell[c]{\includegraphics[width=0.16\linewidth,height=0.098\linewidth]{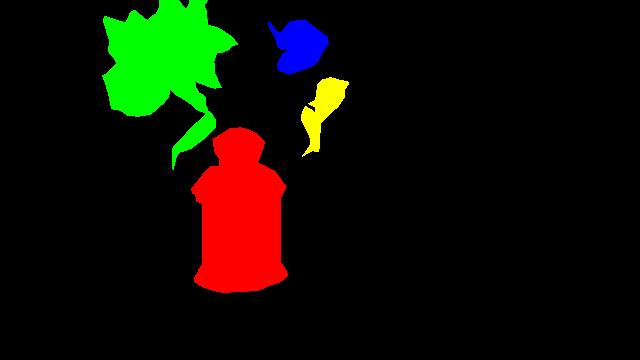}} &
        \makecell[c]{\includegraphics[width=0.16\linewidth,height=0.098\linewidth]{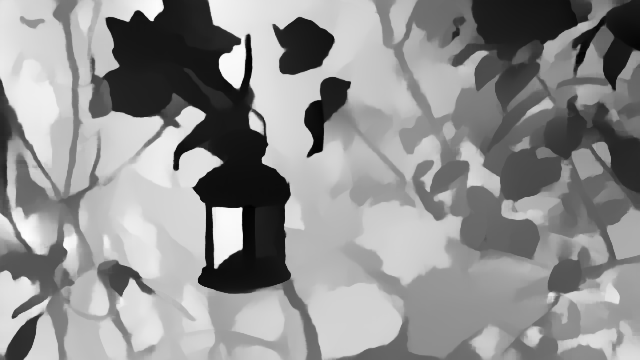}} &
        \makecell[c]{\includegraphics[width=0.16\linewidth,height=0.098\linewidth]{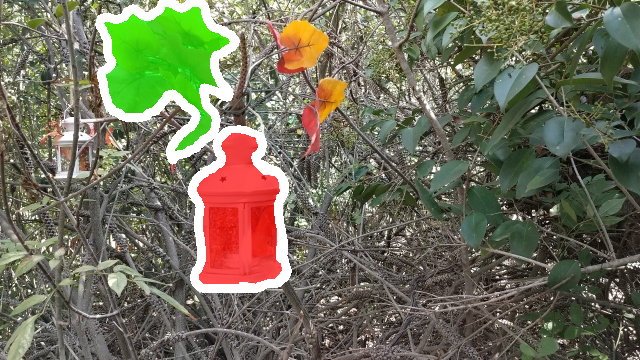}} &
        \makecell[c]{\includegraphics[width=0.16\linewidth,height=0.098\linewidth]{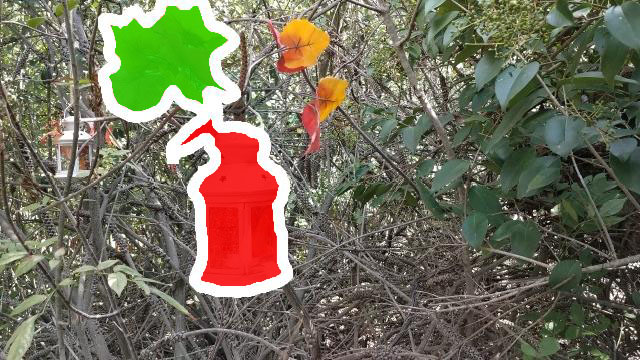}} &
        \makecell[c]{\includegraphics[width=0.16\linewidth,height=0.098\linewidth]{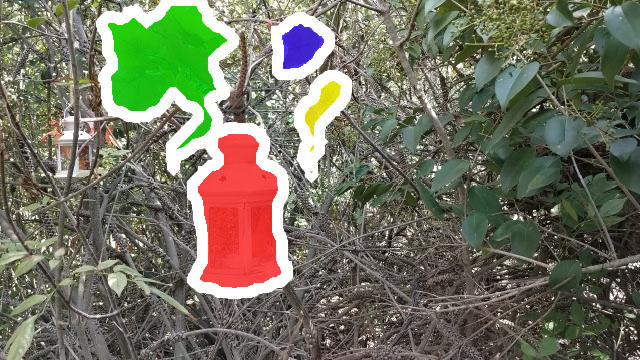}}
        \\
        \makecell[c]{\includegraphics[width=0.16\linewidth,height=0.098\linewidth]{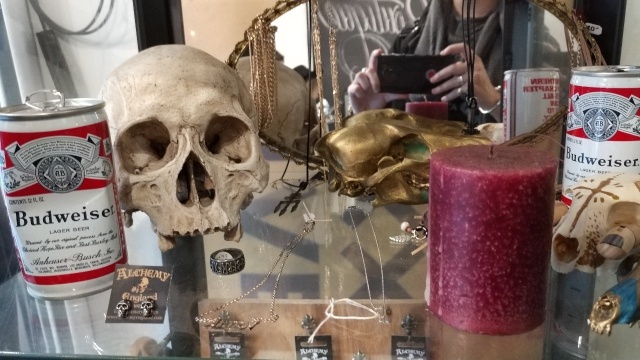}} &
        \makecell[c]{\includegraphics[width=0.16\linewidth,height=0.098\linewidth]{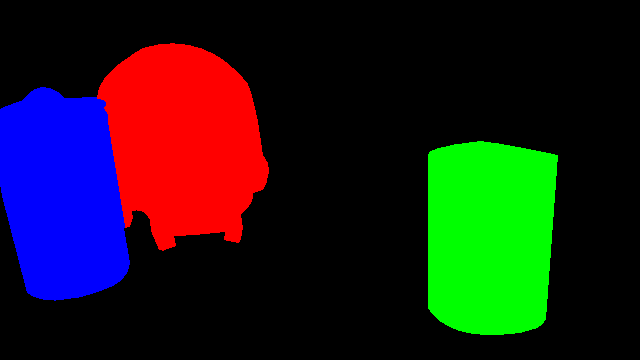}} &
        \makecell[c]{\includegraphics[width=0.16\linewidth,height=0.098\linewidth]{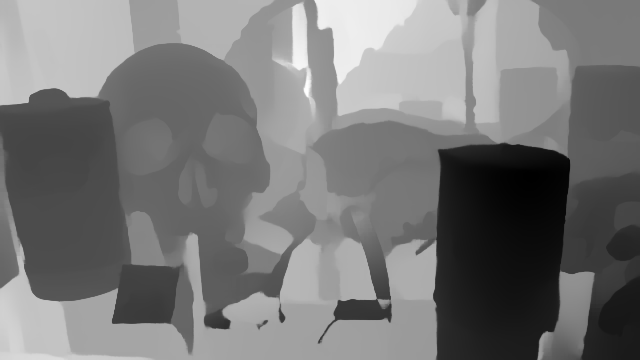}} &
        \makecell[c]{\includegraphics[width=0.16\linewidth,height=0.098\linewidth]{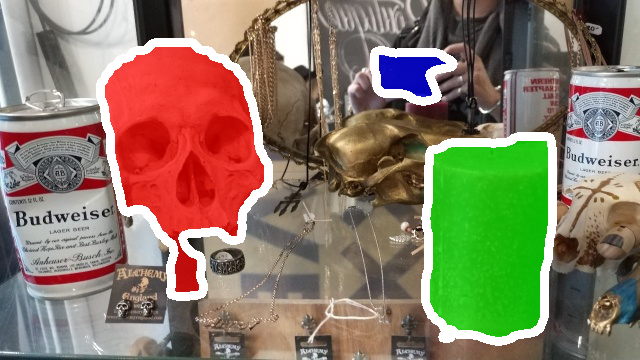}} &
        \makecell[c]{\includegraphics[width=0.16\linewidth,height=0.098\linewidth]{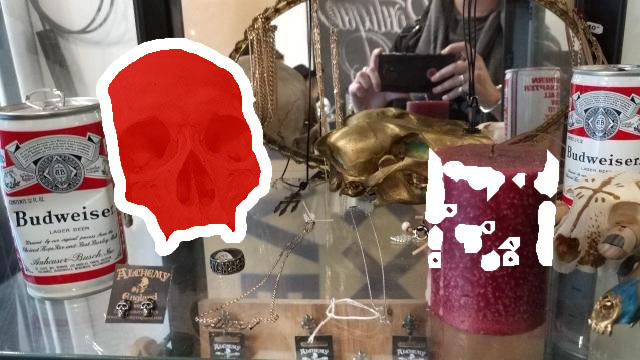}} &
        \makecell[c]{\includegraphics[width=0.16\linewidth,height=0.098\linewidth]{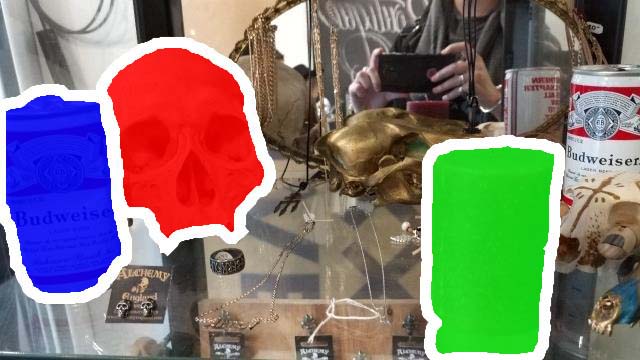}}
        \\
        \makecell[c]{\includegraphics[width=0.16\linewidth,height=0.098\linewidth]{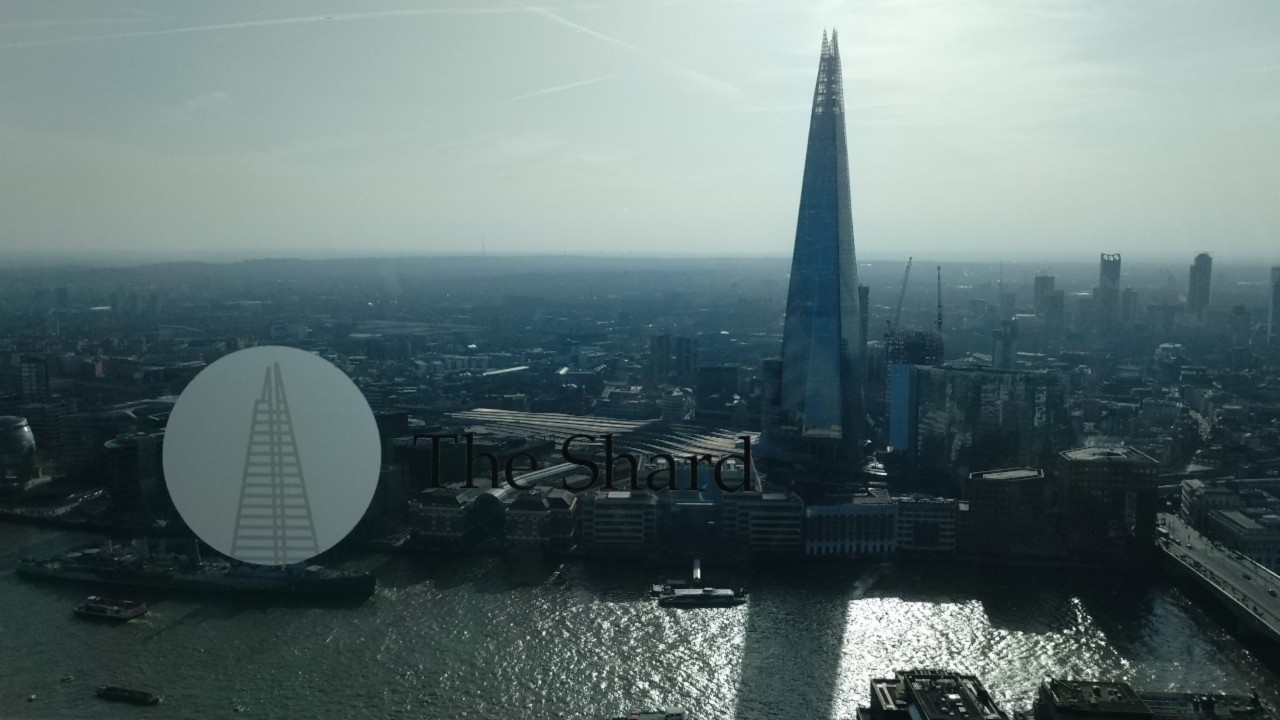}} &
        \makecell[c]{\includegraphics[width=0.16\linewidth,height=0.098\linewidth]{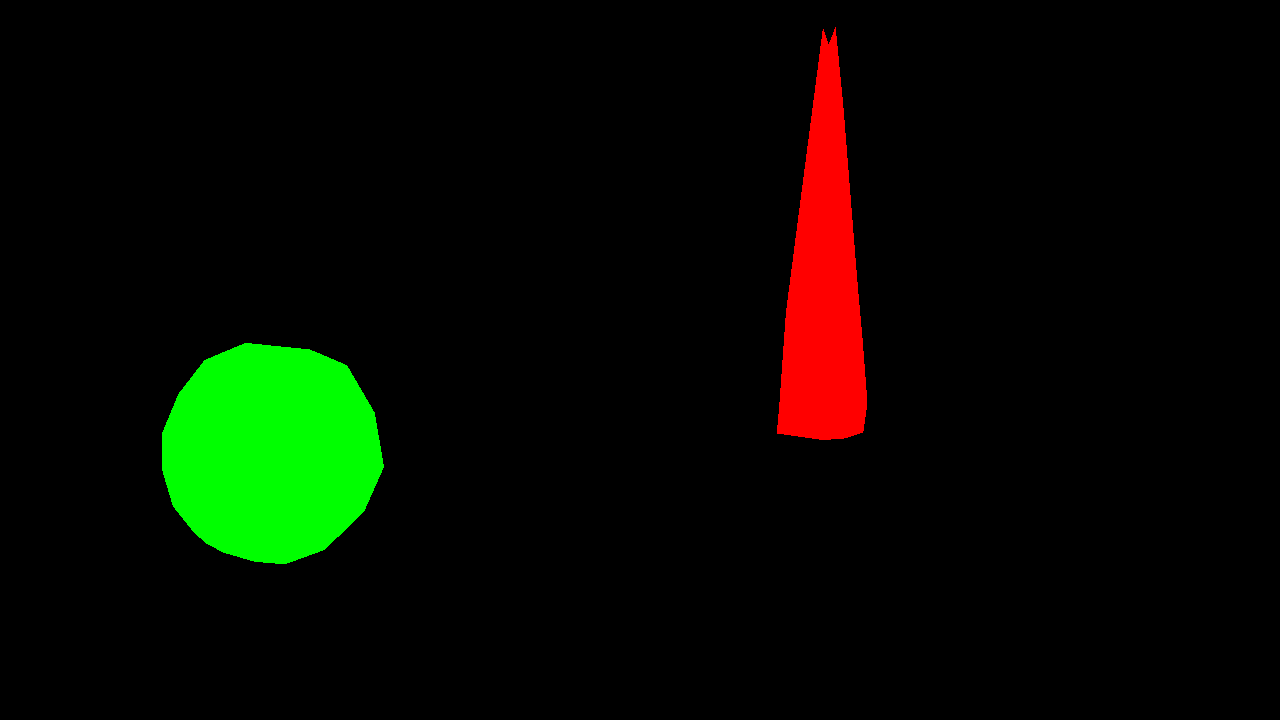}} &
        \makecell[c]{\includegraphics[width=0.16\linewidth,height=0.098\linewidth]{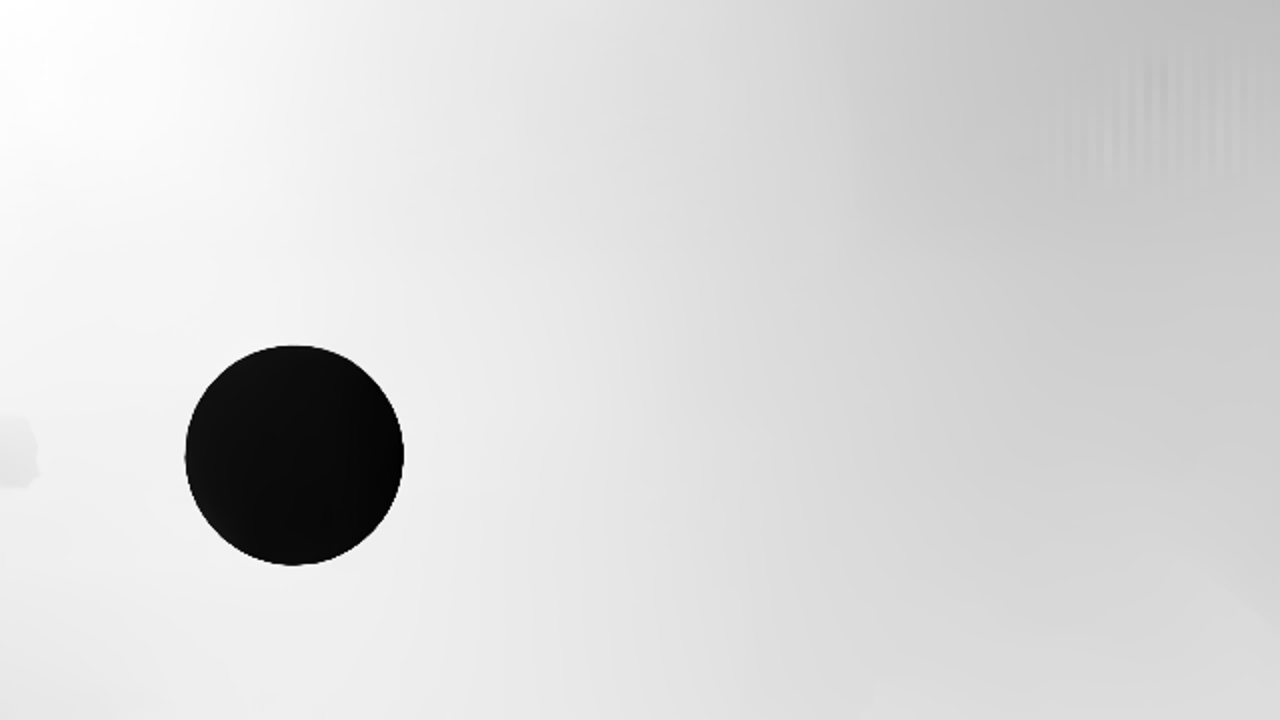}} &
        \makecell[c]{\includegraphics[width=0.16\linewidth,height=0.098\linewidth]{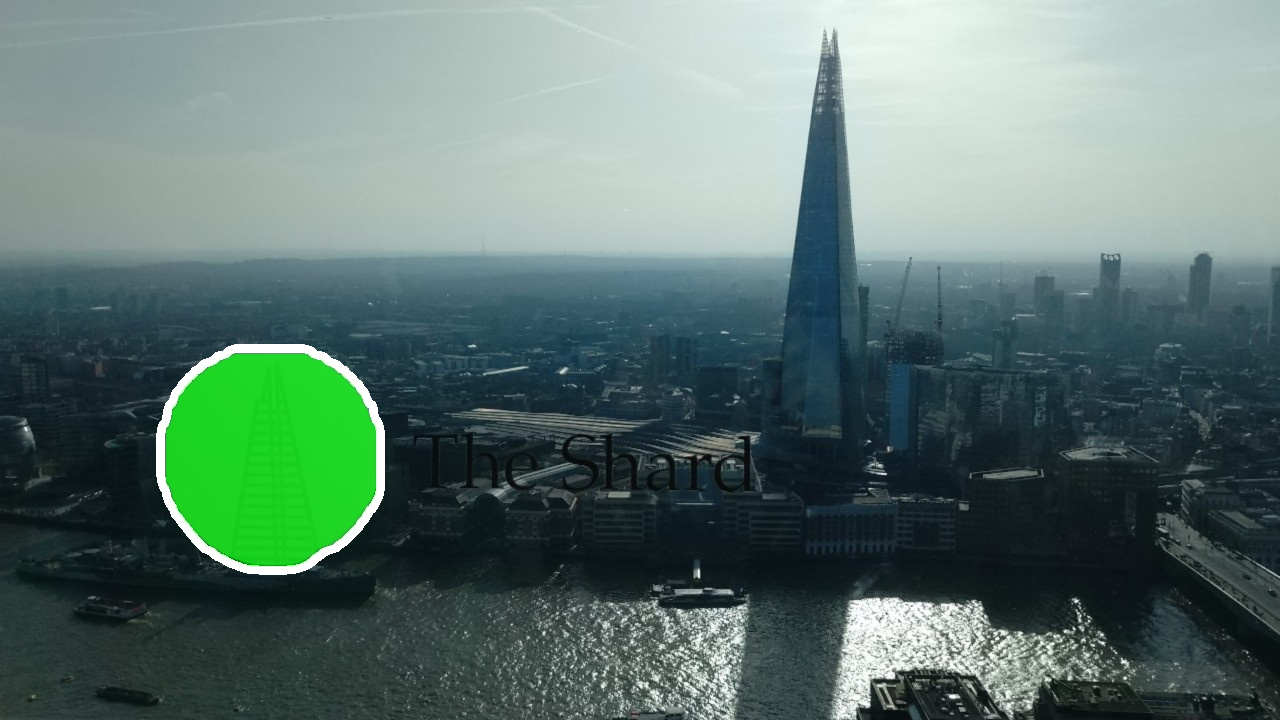}}&
        \makecell[c]{\includegraphics[width=0.16\linewidth,height=0.098\linewidth]{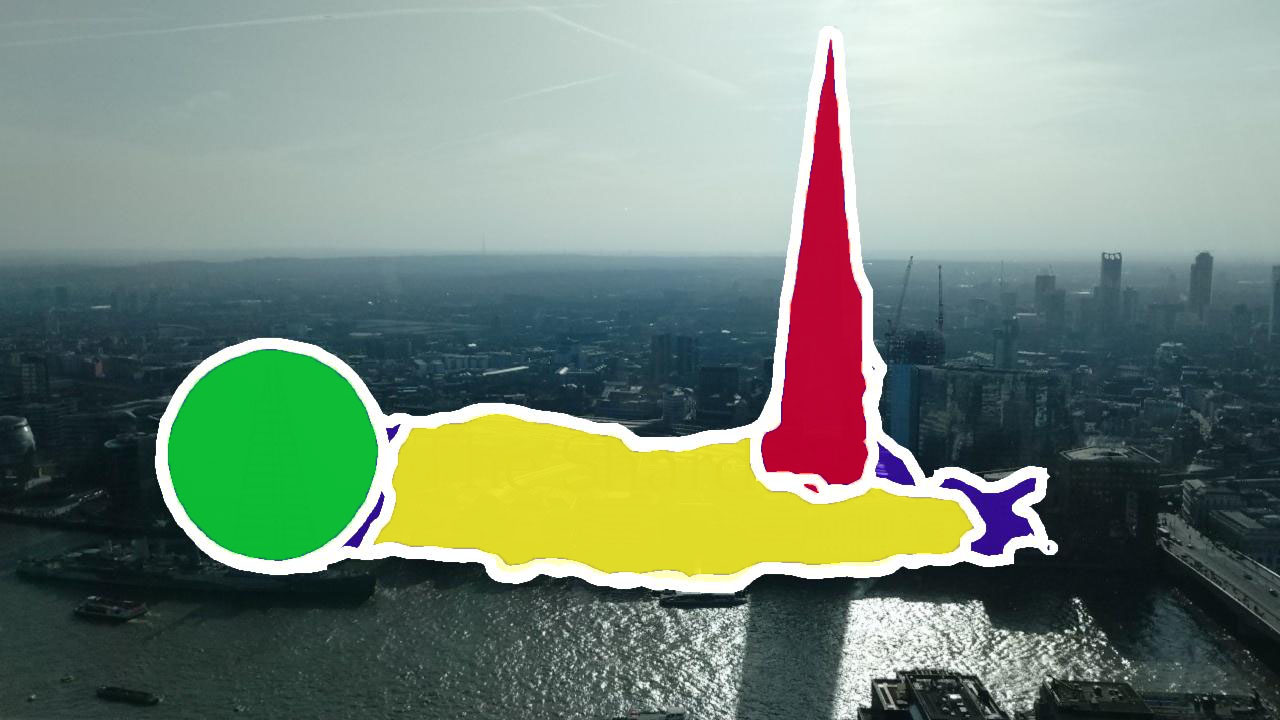}} &
        \makecell[c]{\includegraphics[width=0.16\linewidth,height=0.098\linewidth]{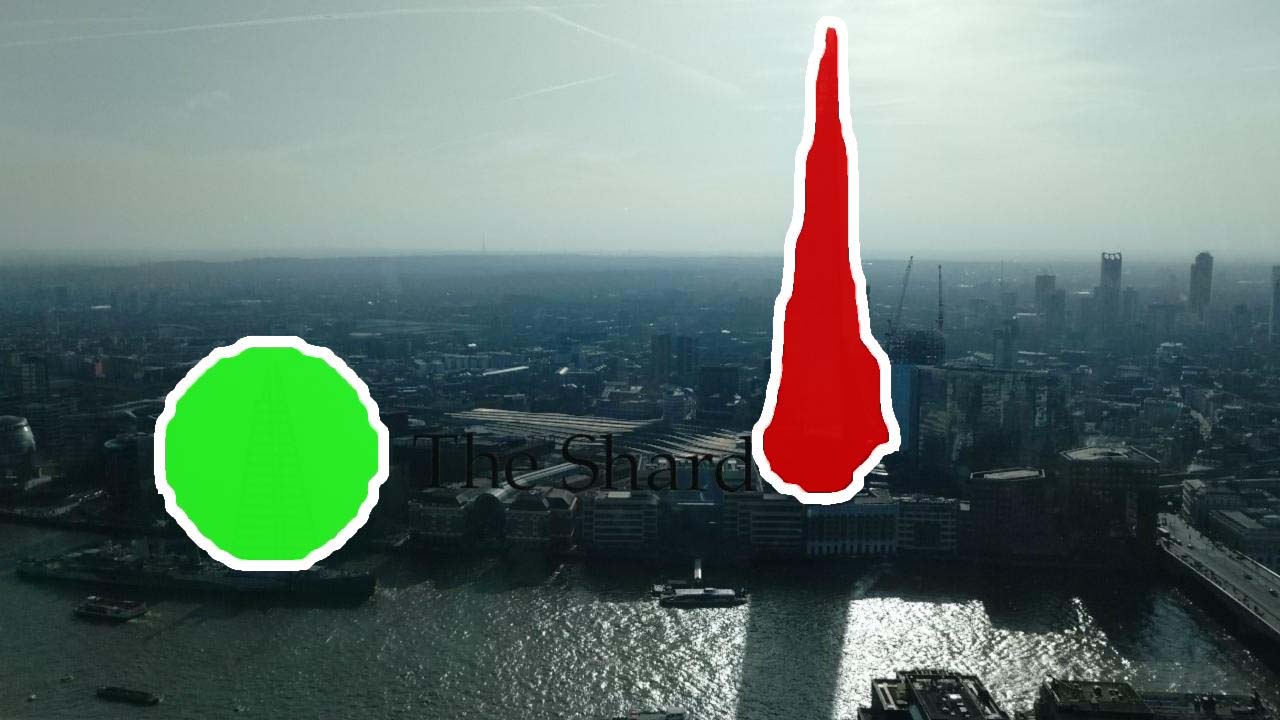}}
        \\
        \makecell[c]{\includegraphics[width=0.16\linewidth,height=0.098\linewidth]{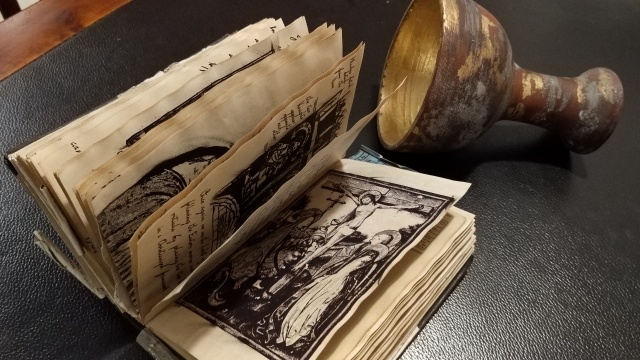}} &
        \makecell[c]{\includegraphics[width=0.16\linewidth,height=0.098\linewidth]{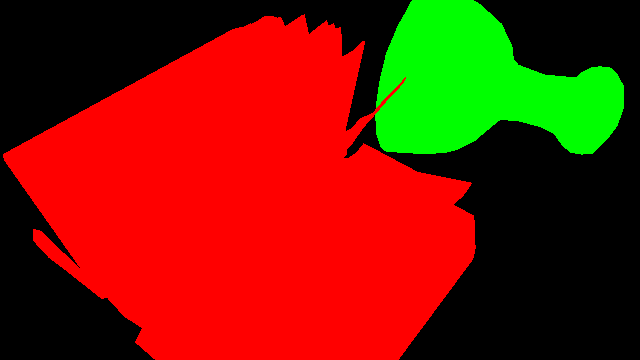}} &
        \makecell[c]{\includegraphics[width=0.16\linewidth,height=0.098\linewidth]{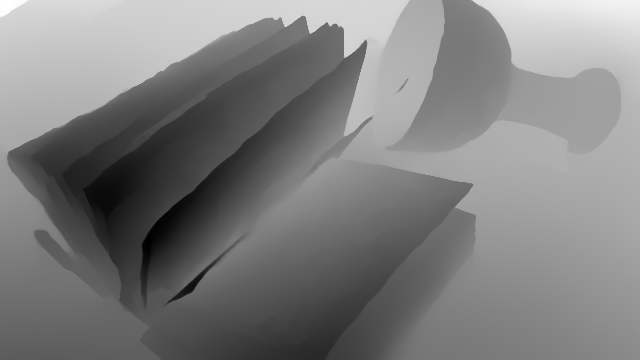}} &
        \makecell[c]{\includegraphics[width=0.16\linewidth,height=0.098\linewidth]{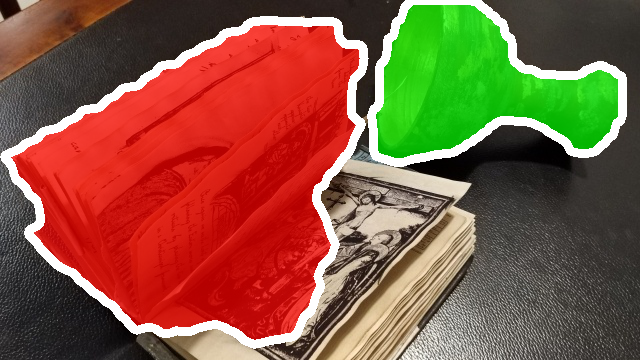}} &
        \makecell[c]{\includegraphics[width=0.16\linewidth,height=0.098\linewidth]{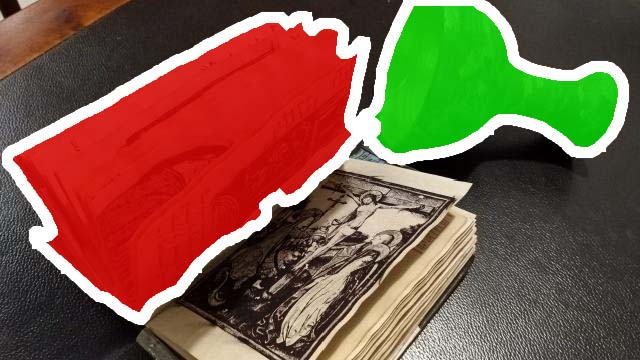}} &
        \makecell[c]{\includegraphics[width=0.16\linewidth,height=0.098\linewidth]{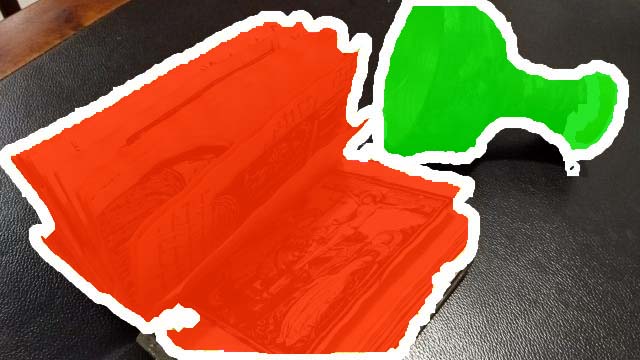}}
        \\
        \makecell[c]{\includegraphics[width=0.16\linewidth,height=0.098\linewidth]{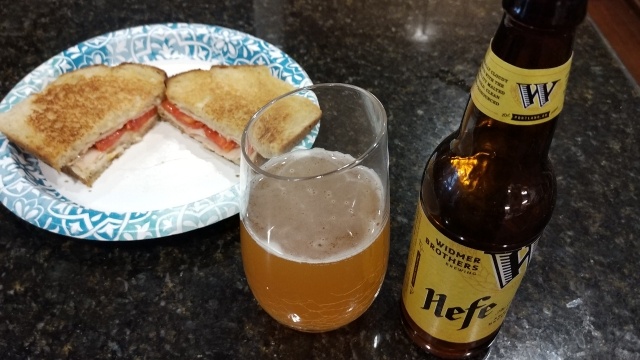}} &
        \makecell[c]{\includegraphics[width=0.16\linewidth,height=0.098\linewidth]{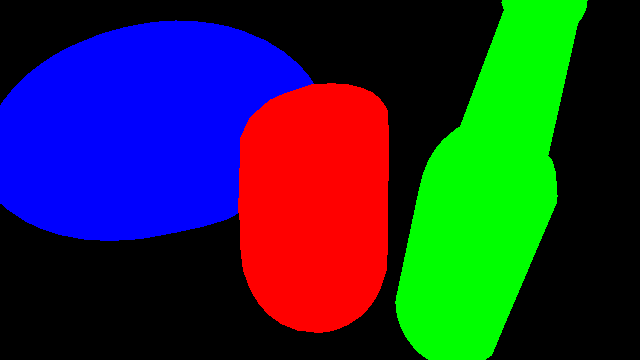}} &
        \makecell[c]{\includegraphics[width=0.16\linewidth,height=0.098\linewidth]{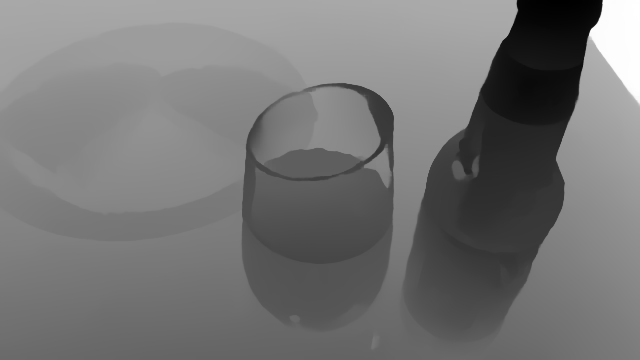}} &
        \makecell[c]{\includegraphics[width=0.16\linewidth,height=0.098\linewidth]{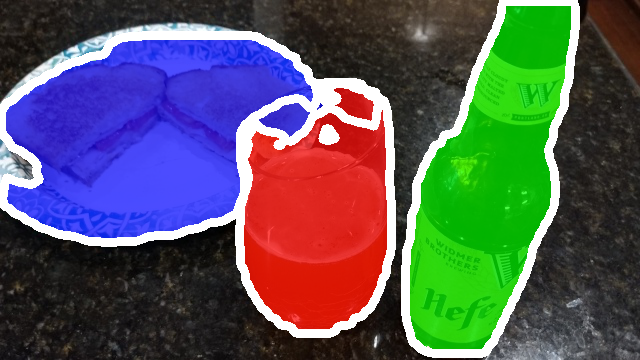}} &
        \makecell[c]{\includegraphics[width=0.16\linewidth,height=0.098\linewidth]{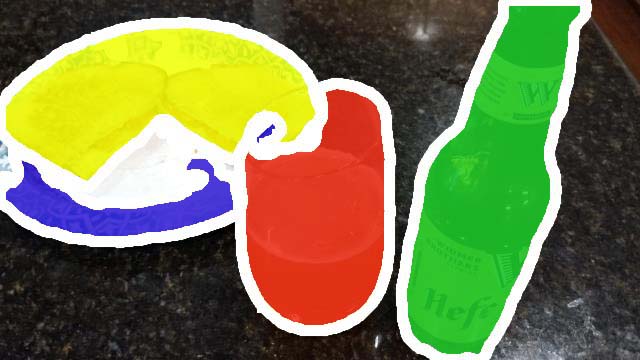}} &
        \makecell[c]{\includegraphics[width=0.16\linewidth,height=0.098\linewidth]{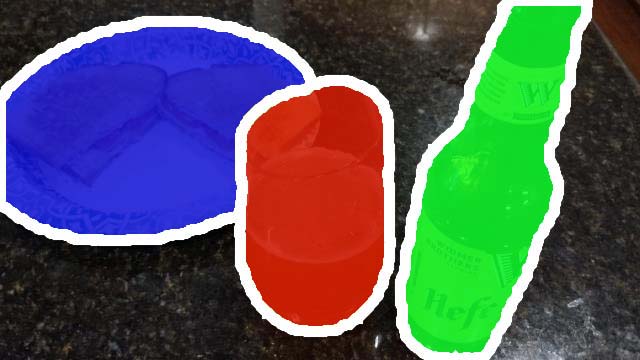}}
        \\
        \makecell[c]{\includegraphics[width=0.16\linewidth,height=0.098\linewidth]{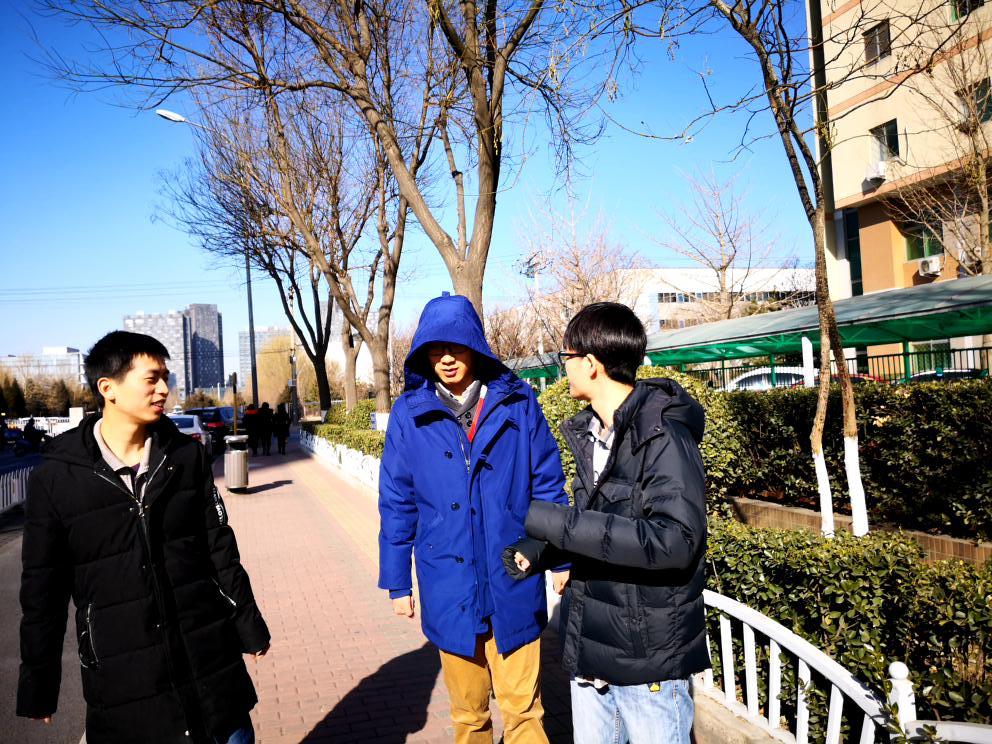}} &
        \makecell[c]{\includegraphics[width=0.16\linewidth,height=0.098\linewidth]{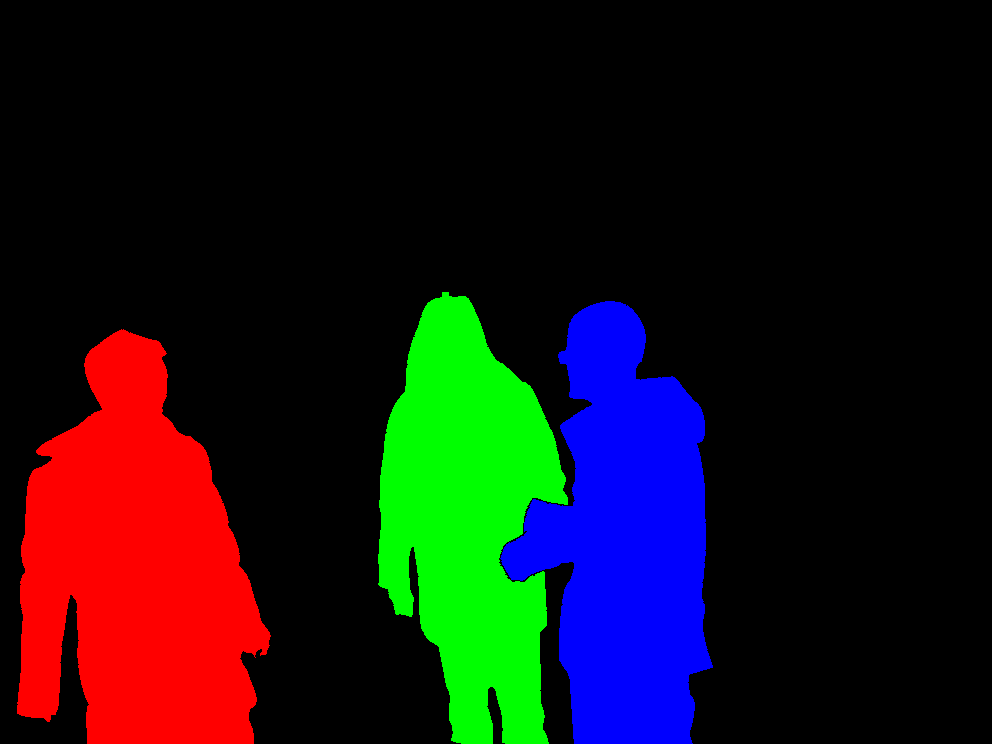}} &
        \makecell[c]{\includegraphics[width=0.16\linewidth,height=0.098\linewidth]{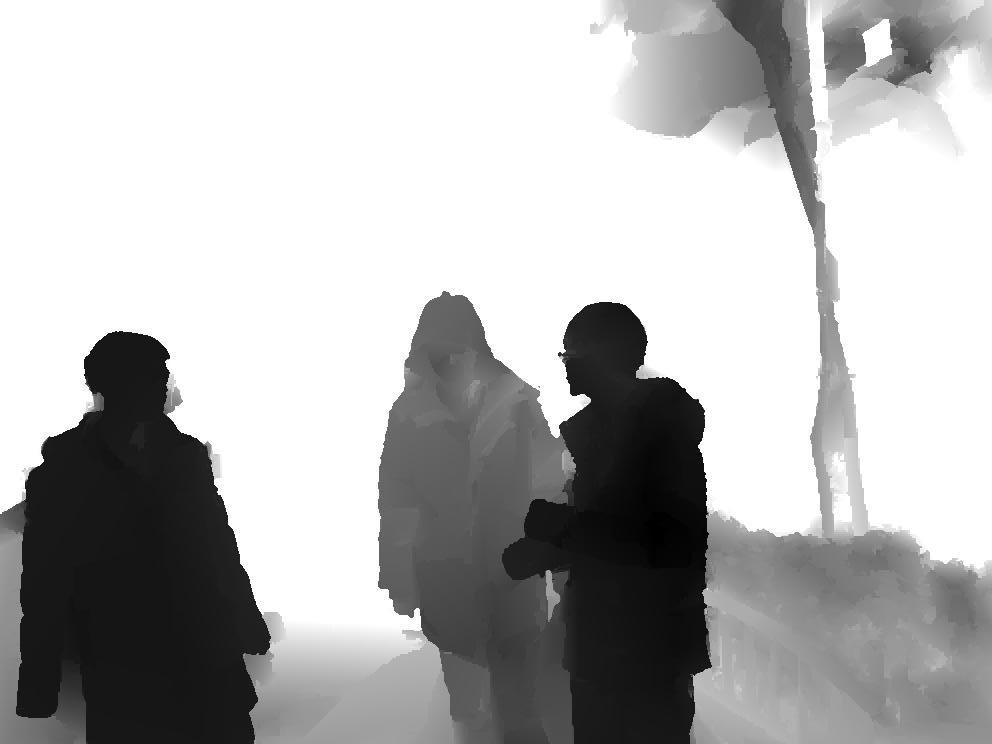}} &
        \makecell[c]{\includegraphics[width=0.16\linewidth,height=0.098\linewidth]{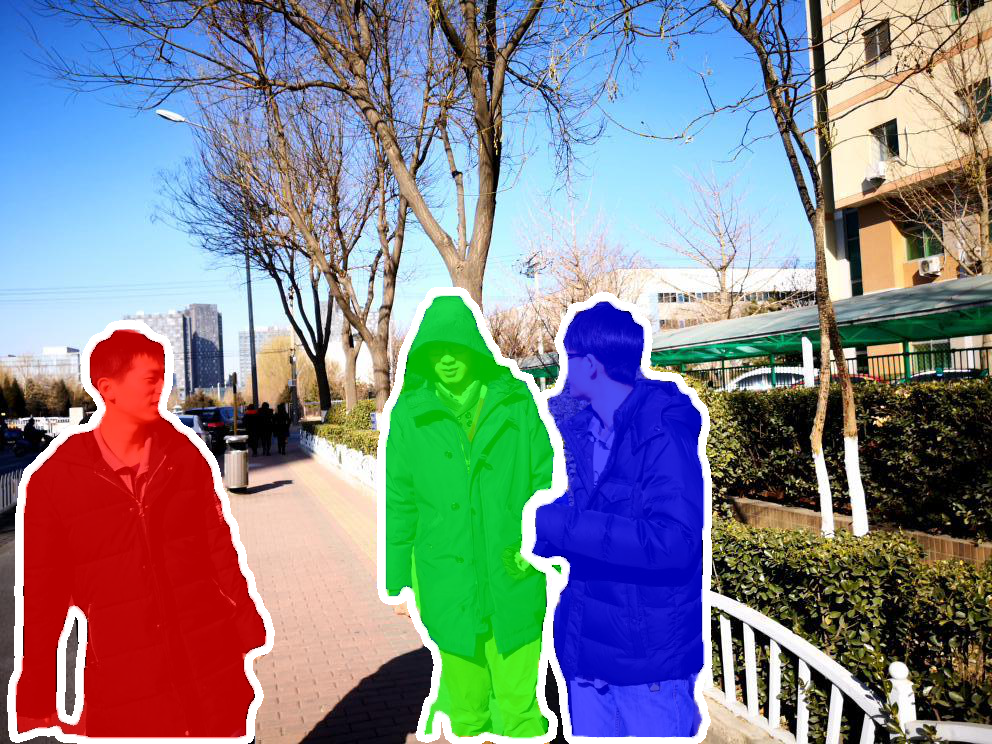}} &
        \makecell[c]{\includegraphics[width=0.16\linewidth,height=0.098\linewidth]{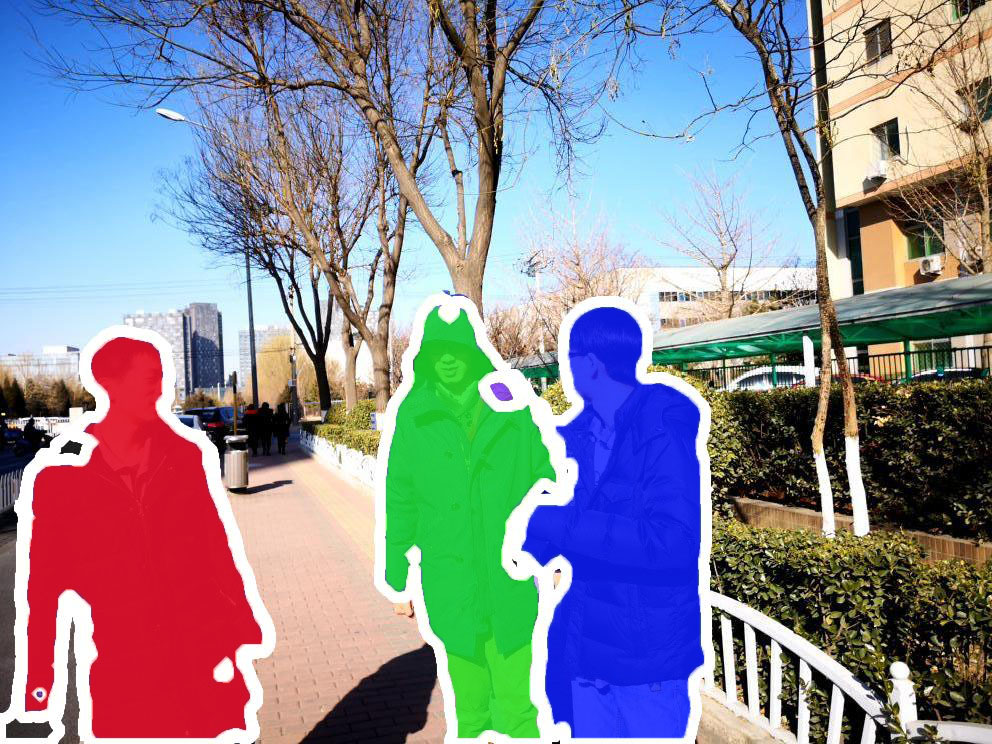}} &
        \makecell[c]{\includegraphics[width=0.16\linewidth,height=0.098\linewidth]{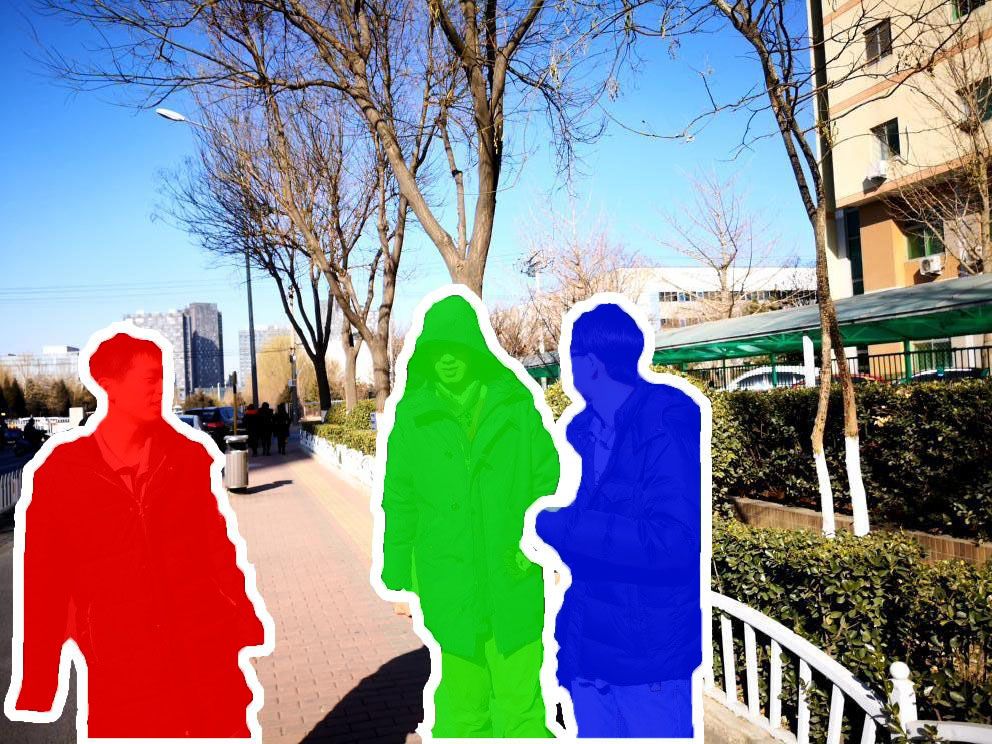}}
        \\
        RGB &
	GT &
	Depth &
	  RDPNet &
	  Mask2Former &
        \textbf{CalibNet}
        \\
    \end{tabular}
    \caption{Visual comparison of~\ourmodel~with representative models on all four RGB-D SIS test sets.}
    \label{fig:visual}
\end{figure*}

\textbf{Quantitative Comparison.}
The prediction results of all compared models are shown in \tabref{tab:SOTA}. 
Our~\ourmodel~consistently outperforms all competitors by a large margin across all three datasets, \eg, 5.0\% AP boost on the COME15K-E test set. Even on the challenging COME15K-H test set, \ourmodel~still reaches a 50.7\% AP score and a 70.4\% AP$_{50}$ score. 
It demonstrates the robustness of our method on various complex scenes.
In addition, we can observe that the performance advantage of the RGB-D cross-modal model on the DSIS test set is not so obvious, \eg, the AP of RRL~\cite{xu2020outdoor} and \ourmodel~on the DSIS test set reaches only 66.1\% and 69.3\% AP scores. This can be attributed to the lower consistency between the salient regions and the depth data in DSIS (refer to \figref{dataset_analysis} (c)), resulting in greater difficulty in exploiting depth information. Conversely, most of the RGB-only GIS models are not required to consider the consistent distribution of the depth maps.
For further comparison, we pick the top three competitive models and retrain them by fusing the depth modal to their frameworks (\ie, OSFormer$^{*}$, Mask2Former$^{*}$, and QueryInst$^{*}$ in \tabref{tab:D-SOTA}). 
For each model, we generally embed a separate backbone in parallel to the original RGB backbone to extract multi-scale depth features. Then, we concatenate multi-level features of two modalities along the channel dimension. Following this, we use a 1$\times$1 convolution together with a group normalization to obtain cross-modal features, subsequently halving the channel dimension at each scale. Other components remain unchanged in the vanilla architecture.
As shown in \tabref{tab:D-SOTA}, Mask2Former$^{*}$ has a 4.4\% AP improvement on the COME15K-E test set after merging depth data, while the performance of OSFormer$^{*}$ is even slightly lower than using the RGB modality only.
Overall, our model still achieves superior performance to other models that have added depth information.
It illustrates the efficient exploitation and integration of the depth modality by the proposed DIK and WSF modules. 
Under the multi-modal architecture, our model is also able to achieve a real-time inference speed of 35.9$fps$.
This result also proves that the depth modality can contribute to improved predictions of instance-level saliency detection, which is in agreement with the conclusion obtained in~\cite{desingh2013depth}.

\textbf{Qualitative Comparison.}
We further intuitively display typical visualization results comparing~\ourmodel~with RDPNet~\cite{wu2021regularized} and Mask2former~\cite{cheng2022masked} evaluated on four RGB-D SIS test sets, COME15K-E, COME15K-H \cite{zhang2021rgb}, DSIS, and SIP test sets \cite{fan2020rethinking}.
As shown in \figref{fig:visual}, our~\ourmodel~is capable of accurately identifying and segmenting the most visually salient instances in a variety of scenes, while suppressing background clutter and other irrelevant targets.
For instance, in the top image of \figref{fig:visual}, which contains two girls and various recreational facilities, \ourmodel~can precisely discriminate them while suppressing the toy vehicle in front of the image. 
Even in the presence of complex backgrounds with cluttered attributes, such as the fourth and fifth rows in~\figref{fig:visual}, the proposed model is still able to accurately detect salient instances.
In the face of changing illumination, especially when the goals have smaller sizes (the sixth row), our method can still handle it.
More importantly, as an instance-level RGB-D model, \ourmodel~is  shown to be effective in handling a range of challenging scenarios even though the structural information of the depth map has a strong misalignment with salient regions (the first and sixth rows). 
It demonstrates the efficiency of our WSF and DSA module in the calibration of cross-modal features.
Compared to the visualizations of RDPNet and Mask2former, \ourmodel~can accurately capture instance edges and produce more intact masks no matter in the case of multi-objects or occlusions, which is also attributed to depth information calibration and enhancement. 


\begin{table}
\begin{center}
\caption{Ablations for the RGB and depth feature fusion in DIK.}
\label{tab:ablation-DIK}
\footnotesize
    \renewcommand{\arraystretch}{1.2}
    \renewcommand{\tabcolsep}{2.7mm}
\begin{tabular}{c|ccc|ccc}
\toprule
\multirow{2}{*}{DIK} & \multicolumn{3}{c|}{COME15K-E}                                  & \multicolumn{3}{c}{COME15K-H}                             \\ \cline{2-7} 
                     & \multicolumn{1}{c}{AP} & \multicolumn{1}{c}{AP$_{50}$} & AP$_{70}$ & \multicolumn{1}{c}{AP} & \multicolumn{1}{c}{AP$_{50}$} & AP$_{70}$ \\ \hline \hline
        Early-fusion &   57.2          &       75.7      &   65.0    &    48.7       &    69.6      &  54.9   \\
\rowcolor[RGB]{235,235,235}
     Later-fusion         &     \textbf{58.0}       &        \textbf{75.8}      &      \textbf{65.6}        &      \textbf{50.7}        &      \textbf{70.4}   &    \textbf{57.3}    \\ \bottomrule
\end{tabular}
\end{center}
\end{table}

\subsection{Ablation Study}\label{ablation}

To verify the effectiveness of the proposed model, we carry out a series of ablations on the COME15K dataset to optimize each component of~\ourmodel. 

\textbf{Modal Fusion in DIK.}
In the DIK module, we attempted to fuse RGB and depth features for generating kernels. 
Instead of using the default design that fuses two modal features after the sigmoid function (later-fusion), we also try to concatenate them first and then share the subsequent operations in DIK (early-fusion). 
\tabref{tab:ablation-DIK} shows that the later-fusion style obtains higher accuracy. It can be explained that the later-fusion design can utilize the structural information and appearance features separately to cover more potential candidates when generating instance-aware kernels.

\begin{table}
\begin{center}
\caption{Comparison of different designs for the shared affinity weight $W_{s}$ in WSF on the COME15K-E test set.}
\label{tab:ablation-share-weight}
\footnotesize
	    \renewcommand{\arraystretch}{1.2}
        \renewcommand{\tabcolsep}{1.8mm}
\begin{tabular}{c|ccc|c|c|c}
\toprule
       Shared Weight          &          AP             &      AP$_{50}$    &      AP$_{70}$   &    Params   &  FLOPs   &  FPS         \\ \hline \hline
     Non-local ($hw$$\times$$hw$)     &      \textbf{57.4}         &   \textbf{75.4}       &    \textbf{65.0}        &  75.9M   &   64.6G   &   29.4  \\ 
\rowcolor[RGB]{235,235,235}
   Ours ($h$$\times$$h$)     &   57.0    &    75.1      &    64.5    &   75.9M  &    \textbf{54.0}G   &  \textbf{39.6}  \\  \bottomrule
\end{tabular}
\end{center}
\end{table}

\textbf{Shared Weight in WSF.}
For WSF module, the shared affinity weight $W_{s}\in\mathbb{R}^{h \times h}$ is designed as the key component to establish strong correlations between RGB and depth modalities and mine mutual information.
To validate the efficiency, we try to adopt the non-local-like pattern~\cite{wang2018non, cong2022cir} in~\ourmodel~ to generate the global shared weight $W_{s}^{g}\in\mathbb{R}^{hw \times hw}$. 
Concretely, the RGB and depth spatial weight maps $W_{r}\in\mathbb{R}^{ h \times w}$ and $W_{d}\in\mathbb{R}^{h \times w}$ are flattened to $L_{r}\in\mathbb{R}^{ 1 \times hw}$ and $L_{d}\in\mathbb{R}^{1 \times hw}$. 
Then, $L_{r}$ and $L_{d}$ are also operated by matrix multiplication to obtain $W_{s}^{g}$. 
For more efficient comparison, we use only the single-scale feature ($T_{c}^{3}$ and $T_{d}^{3}$) as input to the mask branch.
As illustrated in \tabref{tab:ablation-share-weight}, the FLOPs of the model with the non-local shared weight increase considerably, while the improvement in accuracy is limited. 
The FLOPs boost even to 103$G$ when adding the low-level features ($C_{2}$,$D_{2}$) in WSF.
Besides, the full-size weighting map also imposes a burden on subsequent operations, resulting in a substantial drop in the FPS value at the inference time. 
In a nutshell, the default shared affinity weight $W_{s}$ makes \ourmodel~more flexible.

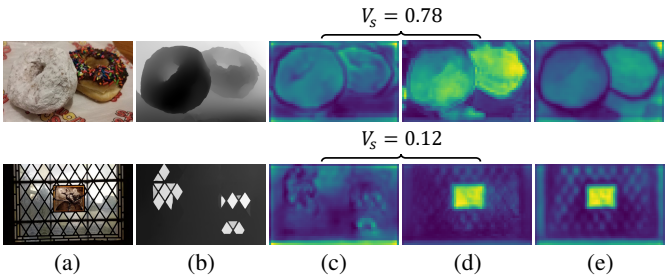
\begin{figure}[t]
\centering
    \begin{overpic}[width=\linewidth]{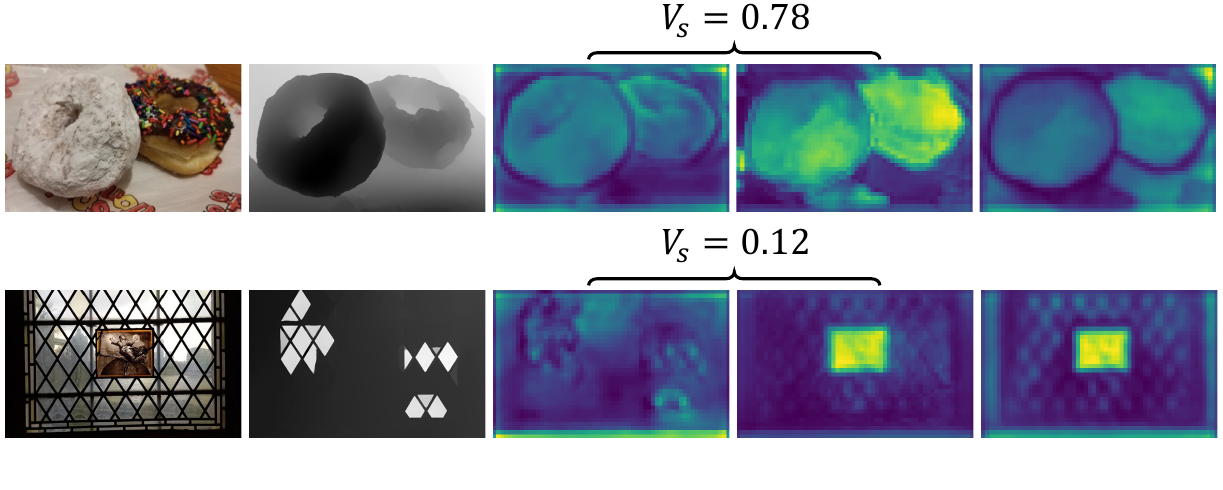}
        \put(8,0.5){\small{(a)}}
        \put(28,0.5){\small{(b)}}
        \put(48,0.5){\small{(c)}}
        \put(68,0.5){\small{(d)}}
        \put(88,0.5){\small{(e)}}
    \end{overpic}
	\caption{Visualization of attention maps in the mask branch. The top is a higher similarity case and the bottom is a lower similarity case. (a) Input image; (b) Depth map; (c) Depth feature input to DSA; (d) RGB feature input to DSA; (e) Generated mask feature.}\label{DSA-feature}
\end{figure}

\begin{table}
\begin{center}
\caption{Ablations for the effect of shared DSA in~\ourmodel.}
\label{tab:ablation-DSA}
\footnotesize
    \renewcommand{\arraystretch}{1.2}
    \renewcommand{\tabcolsep}{2.8mm}
\begin{tabular}{c|ccc|ccc}
\toprule
\multirow{2}{*}{DSA} & \multicolumn{3}{c|}{COME15K-E}                                  & \multicolumn{3}{c}{COME15K-H}                             \\ \cline{2-7} 
                     & \multicolumn{1}{c}{AP} & \multicolumn{1}{c}{AP$_{50}$} & AP$_{70}$ & \multicolumn{1}{c}{AP} & \multicolumn{1}{c}{AP$_{50}$} & AP$_{70}$ \\ \hline \hline
        Shared      &   57.0          &       75.6       &   64.2      &    50.3       &       70.4        &   56.7   \\ 
\rowcolor[RGB]{235,235,235}
       w/o Shared              &     \textbf{58.0}       &        \textbf{75.8}      &      \textbf{65.6}        &      \textbf{50.7}        &      \textbf{70.4}   &    \textbf{57.3}    \\ \bottomrule
\end{tabular}
\end{center}
\vspace{-5pt}
\end{table}

\textbf{Shared Settings for DSA.}
\ourmodel~embeds the DSA module before DIK and WSF to distill depth features for better feature consistency. 
In \tabref{tab:ablation-DSA}, we compare the settings of using a shared DSA versus non-shared DSAs in the mask and kernel branches. 
It indicates that~\ourmodel~with shared DSA performs relatively worse because the demand for RGB and depth feature discretization is different in the mask and kernel branches. 
Therefore, our framework is appropriate for using non-shared DSA modules in different branches. 
We also visualize the input features of RGB and depth in DSA and the corresponding similarity scores in \figref{DSA-feature}. 
We can see that the visual difference between the two modal features is consistent with the quantified scores.

\textbf{Multiscale Fusion in Mask Branch.} 
The proposed WSF is responsible for fusing cross-modal features in the mask branch. 
It is also essential to integrate multiscale cross-level features from RGB and depth encoders to obtain the unified mask feature. 
Hence, we conduct an ablation experiment to combine different layers in the mask branch so as to evaluate the effectiveness of our model. 
The results of various combinations are shown in \tabref{tab:ablation-WSF}.
The first two rows demonstrate the importance of low-level features. Besides, adding more transformer layers has no remarkable effect while bringing more extra parameters. 
Considering the trade-off between accuracy and efficiency, we finally adopt the second layer in the CNN backbone ($C_{2}$, $D_{2}$) and the third layer after the transformer ($T_{c}^{3}$, $T_{d}^{3}$) for multiscale fusion. 
The mask feature achieved by the mask branch is also visualized in the last column of \figref{DSA-feature}.

\begin{table}
\begin{center}
\caption{Comparison of multiscale feature fusion in the mask branch on the COME15K-E test set.}
\label{tab:ablation-WSF}
\scriptsize
	    \renewcommand{\arraystretch}{1.5}
        \renewcommand{\tabcolsep}{0.8mm}
\begin{tabular}{l|ccc|c|c|c}
\toprule
	    Combinations                 &          AP             &      AP$_{50}$    &      AP$_{70}$   &    Params   &  FLOPs   &  FPS         \\ \hline \hline
 ($T_{c}^{3}$,$T_{d}^{3}$)               &         57.0         &      75.1           &        64.5          &  \textbf{75.9}M   &  \textbf{54.0}G   &   \textbf{39.6}  \\ 
   	    \rowcolor[RGB]{235,235,235}
 ($C_{2}$,$D_{2}$); ($T_{c}^{3}$,$T_{d}^{3}$)      &      \textbf{58.0}       &        \textbf{75.8}      &      \textbf{65.6}          &     78.9M     &     90.2G    &    35.9      \\
 ($C_{2}$,$D_{2}$); ($T_{c}^{3}$,$T_{d}^{3}$); ($T_{c}^{4}$,$T_{d}^{4}$)   &      57.3          &      75.6         &       65.0       &   81.9M   &    91.8G &   35.2 \\ 
($T_{c}^{3}$,$T_{d}^{3}$); ($T_{c}^{4}$,$T_{d}^{4}$); ($T_{c}^{5}$,$T_{d}^{5}$)      &    57.1       &      75.4          &        64.7        &  81.9M   &  56.0G &   35.8     \\
 ($C_{2}$,$D_{2}$); ($T_{c}^{3}$,$T_{d}^{3}$); ($T_{c}^{4}$,$T_{d}^{4}$); ($T_{c}^{5}$,$T_{d}^{5}$)      &   57.5    &    \textbf{75.8}       &    64.8    &   84.8M &  92.2G &  34.3  \\  \bottomrule
\end{tabular}
\end{center}
\end{table}

\begin{table}[t!]
\begin{center}
\caption{Ablations for different numbers of instance-aware kernels.}
\label{tab:ablation-kernels}
\footnotesize
    \renewcommand{\arraystretch}{1.2}
    \renewcommand{\tabcolsep}{1.6mm}
\begin{tabular}{c|ccc|ccc|c|c}
\toprule
\multirow{2}{*}{$N$} & \multicolumn{3}{c|}{COME15K-E}               & \multicolumn{3}{c|}{COME15K-H}   &  \multirow{2}{*}{Params}  &    \multirow{2}{*}{GFLOPs}                    \\ \cline{2-7} 
                & \multicolumn{1}{c}{AP} & \multicolumn{1}{c}{AP$_{50}$} & AP$_{70}$ & \multicolumn{1}{c}{AP} & \multicolumn{1}{c}{AP$_{50}$} & AP$_{70}$   &   \multicolumn{1}{c|}{}  &   \multicolumn{1}{c}{} \\ \hline \hline
        10    &     56.5     &    75.0         &   64.0    &  48.5        &  69.2       &  54.3    & \textbf{78.7}M    &  \textbf{89.5}  \\
     20      &      57.2    &   75.2   &   64.6    &   50.0     &    70.3    &    56.1    &  78.8M      &   89.7  \\ 
\rowcolor[RGB]{235,235,235}
50      &     58.0     &      75.8        &    \textbf{65.6}     &   \textbf{50.7}        &   70.4      &  \textbf{57.3}   &  78.9M    &  90.2  \\
 100   &    \textbf{58.1}      &      \textbf{76.3}     &     \textbf{65.6}   &   50.3       &   \textbf{70.9}      &   57.2   &   79.2M   & 91.2  \\  \bottomrule
\end{tabular}
\end{center}
\end{table}

\textbf{Number of Kernels.}
The appropriate number of instance-aware kernels can adequately locate potentially salient instances and balance the accuracy and efficiency of the proposed~\ourmodel. 
To investigate the effect of the number of kernels, we perform an ablation experiment with different numbers of kernels $N$ (10, 20, 50, and 100).
As shown in \tabref{tab:ablation-kernels}, the performance of~\ourmodel~improves consistently with the increase in the number of kernels.
When $N$ is greater than 50, the improvement in accuracy becomes marginal, and even the AP value drops slightly on the COME15K-H test set.
Trading off the accuracy and efficiency of our model, the number of instance-aware kernels $N$ in~\ourmodel~is set to 50.

\textbf{Number of Transformer Layers.}
The integration of deformable transformer layers into both RGB and depth encoders in~\ourmodel~is aimed at enhancing the capture of global information. 
To optimize the performance of our model, we experimented with different combinations of the number of layers on the RGB and depth encoders. 
As illustrated in \tabref{tab:ablation-trans-layer}, the first row indicates that it is not appropriate to add transformer layers to only the RGB modality.
When the number of layers in RGB and depth encoders is 3 and 3, respectively, \ourmodel~achieves promising results in both COME15K-E and COME15K-H test sets~\cite{zhang2021rgb}.
Although increasing the number of transformer layers in the RGB encoder to 6 leads to higher AP values on the COME15K-E test set, the additional parameters of the model could have a negative impact on the multimodal architecture.
As a result, the default number of transformer layers in RGB and depth encoders is set to 3 for both modalities.

\begin{table}
\begin{center}
\caption{Performance of~\ourmodel~with different combinations of the number of transformer layers in RGB and depth encoders.}
\label{tab:ablation-trans-layer}
\footnotesize
    \renewcommand{\arraystretch}{1.2}
    \renewcommand{\tabcolsep}{1.7mm}
\begin{tabular}{cc|ccc|ccc|c}
\toprule
 \multicolumn{2}{c|}{Number}    & \multicolumn{3}{c|}{COME15K-E}               & \multicolumn{3}{c|}{COME15K-H}   &  \multirow{2}{*}{Params}              \\ \cline{1-8} 
   \multicolumn{1}{c}{RGB}       &    \multicolumn{1}{c|}{Depth}     & \multicolumn{1}{c}{AP} & \multicolumn{1}{c}{AP$_{50}$} & AP$_{70}$ & \multicolumn{1}{c}{AP} & \multicolumn{1}{c}{AP$_{50}$} & AP$_{70}$   &   \multicolumn{1}{c}{}  \\ \hline \hline
   3   &     0    &     57.2    &     75.4        &  64.9  &    49.6     &   70.5     &  56.4   &    76.7M   \\
   3   &     1     &   57.1     &      75.7       & 65.0  &    49.0     &   69.9    &   55.3  &   77.4M    \\
\rowcolor[RGB]{235,235,235}
   3   &    3    &     58.0     &      75.8        &    65.6     &   \textbf{50.7}    &   70.4      &  \textbf{57.3}   &  78.9M    \\
   6   &    1    &     55.4   &   74.6      &  63.2  &   47.2      &    69.0    &  52.9   &   79.6M    \\
   6  &    3     &  \textbf{58.2}  &  \textbf{76.3}  &  \textbf{65.8}  &  50.6   &     \textbf{70.9}   &  57.0   &   81.1M    \\
   6   &    6     &     58.0   &     76.1        &  65.3  &     50.4    &  70.9      &   56.8  &    83.3M   \\  \bottomrule
\end{tabular}
\end{center}
\end{table}

\textbf{Comparison with Other Fusion Strategies.}
We further compare our proposed WSF with other competitive feature interaction strategies in \tabref{tab:ablation-fusion} to validate the effectiveness of WSF. 
The first row stands for replacing the shared affinity weight with the output after a cascaded fusion and channel compression.
The RMFF~\cite{huang2022middle} module calculates middle-level features’ total, shared, and differential information to identify redundant and complementary
relationships.
In contrast, our proposed WSF module uses spatial attention weights to generate a shared affinity map, which makes better use of the structural information in depth maps, allowing for the calibration
and merging of cross-modal information.
As illustrated in \tabref{tab:ablation-fusion}, WSF achieves the best performance among all the strategies, which proves the superiority of WSF.

\begin{table}
\begin{center}
\caption{Comparion of WSF and other fusion strategies.}
\label{tab:ablation-fusion}
    \footnotesize
    \renewcommand{\arraystretch}{1.2}
    \renewcommand{\tabcolsep}{2.2mm}
\begin{tabular}{c|ccc|ccc}
\hline
\multirow{2}{*}{Interactions} & \multicolumn{3}{c|}{COME15K-N}                                  & \multicolumn{3}{c}{COME15K-D}                             \\ \cline{2-7} 
                     & \multicolumn{1}{c}{AP} & \multicolumn{1}{c}{AP$_{50}$} & AP$_{70}$ & \multicolumn{1}{c}{AP} & \multicolumn{1}{c}{AP$_{50}$} & AP$_{70}$ \\ \hline \hline
     Cascade fusion &      57.1     &     75.4      &    64.7   &    49.8       &     69.9     &   56.1   \\
     RMFF~\cite{huang2022middle}     &     57.6     &     75.6      &    65.1      &     49.9      &    70.1     &    55.9   \\
     \rowcolor[RGB]{235,235,235}
          WSF (Ours)    &     \textbf{58.0}       &        \textbf{75.8}      &      \textbf{65.6}        &      \textbf{50.7}        &      \textbf{70.4}   &    \textbf{57.3}    \\ \hline
\end{tabular}
\end{center}
\end{table}

\begin{table}[t!]
\begin{center}
\caption{Ablations for contributions of the objectness loss $\mathcal{L}_{obj}$ and the auxiliary loss $\mathcal{L}_{bin}$ on COME15K test set.}
\label{tab:ablation-loss}
\footnotesize
\renewcommand{\arraystretch}{1.2}
\renewcommand{\tabcolsep}{2.5mm}
\begin{tabular}{cc|ccc|ccc}
\toprule
 \multicolumn{2}{c|}{Loss Items}    & \multicolumn{3}{c|}{COME15K-E}          & \multicolumn{3}{c}{COME15K-H}   \\ \cline{1-8} 
   \multicolumn{1}{c}{$\mathcal{L}_{obj}$}       &    \multicolumn{1}{c|}{$\mathcal{L}_{bin}$}     & \multicolumn{1}{c}{AP} & \multicolumn{1}{c}{AP$_{50}$} & AP$_{70}$ & \multicolumn{1}{c}{AP} & \multicolumn{1}{c}{AP$_{50}$} & AP$_{70}$     \\ \hline \hline
      &     \checkmark    &     56.7    &     75.4        &  64.7  &    49.7     &   70.5     &  56.4    \\
   \checkmark   &          &   57.8     &      \textbf{76.1}       & 65.4  &    50.2     &   \textbf{70.9}    &   57.1     \\
\rowcolor[RGB]{235,235,235}
   \checkmark   &    \checkmark    &     \textbf{58.0}     &      75.8        &    \textbf{65.6}     &   \textbf{50.7}    &   70.4      &  \textbf{57.3}      \\  \bottomrule
\end{tabular}
\end{center}
\end{table}

\textbf{Effect of Loss Items.} The proposed~\ourmodel~employs an IoU-aware objectness loss to separate instances and an auxiliary loss to perform auxiliary supervision at the regional level. We conduct an ablation study to further investigate the contribution of each loss item. As illustrated in~\tabref{tab:ablation-loss}, both the objectness loss and the auxiliary loss contribute positively to the model performance, \eg, bringing approximately 1.3\% and 0.2\% AP improvement on the COME15K-E test set.

\begin{table}[t!]
\begin{center}
\caption{Ablation studies for each component of~\ourmodel~on the COME15K-E test set.}
\label{tab:ablation}
\footnotesize
\renewcommand{\arraystretch}{1.2}
\renewcommand{\tabcolsep}{3.2mm}
\begin{tabular}{c|c|c|c|ccc}
\toprule
Input   &  DIK      &    WSF     &    DSA    &       AP     &      AP$_{50}$    &      AP$_{70}$    \\ \hline \hline
RGB     &           &            &           &      52.9    &      72.8         &      60.5         \\  \hline
\multirow{7}{*}{RGB-D}&       &      &       &      55.2    &      74.1         &       62.9        \\
        & \checkmark  &         &            &      56.5    &      75.3         &      64.0       \\ 
        &           &  \checkmark &          &      56.7    &      75.1         &   64.1     \\ 
                &        &        &  \checkmark    &     55.8    &      75.4      &   64.3  \\  
        &  \checkmark &    &  \checkmark     &      56.9    &      75.3         &   64.9  \\ 
        &     &  \checkmark   &  \checkmark  &      57.3    &      75.6         &    64.7   \\ 
        &   \checkmark &  \checkmark   &     &      57.5    &      75.6         &       65.1       \\
\rowcolor[RGB]{235,235,235} 
        & \checkmark & \checkmark & \checkmark & \textbf{58.0} & \textbf{75.8}  &  \textbf{65.6}          \\   \bottomrule
\end{tabular}
\end{center}
\end{table}

\textbf{Effectiveness of Each Component.} 
To optimize the performance of our model, we conduct an ablation study to thoroughly validate the effectiveness of each component in~\ourmodel, including the dynamic interactive kernel (DIK) module, the weight-sharing fusion (WSF), and the depth similarity assessment (DSA) module. 
For the RGB baseline model, we use only the RGB modality to obtain the instance-aware kernel and mask feature without using any of our proposed parts. 
In the RGB-D input case, as shown in the second row of \tabref{tab:ablation}, We simply add RGB and depth features and fuse them using 1$\times$1 convolution.
It indicates that employing the depth modality brings a 2.3\% AP improvement thanks to the proposed cross-modal dual-branch framework.
As illustrated in the next three rows, each component has a positive impact on the performance of~\ourmodel~by efficient integration of deep features, especially the contribution of DIK and WSF. 
Gathering all parts, our proposed~\ourmodel~can obtain better performance compared to other variants.

\textbf{Performance of Different Backbones.}
%
To exploit the potential of our model, we adopt several backbone networks with ImageNet~\cite{deng2009imagenet} pre-trained weights to train~\ourmodel, including ResNet-50~\cite{he2016deep}, ResNet-101~\cite{he2016deep}, Swin-T~\cite{liu2021swin}, PVTv2-B2-Li~\cite{wang2022pvt} and P2T-Large~\cite{wu2022p2t}. As shown~\tabref{tab:ablation-backbone}, \ourmodel~can achieve 58.0\% AP values based on the default ResNet-50 backbone. With a stronger backbone network (\ie, P2T-Large), our model further reaches 61.8\% AP on the COME15K-E test set. It indicates that the performance of our model still has considerable scalability.

\begin{table}
\begin{center}
\caption{Performance of~\ourmodel~with different backbones.}
\label{tab:ablation-backbone}
    \footnotesize
    \renewcommand{\arraystretch}{1.2}
    \renewcommand{\tabcolsep}{2.2mm}
\begin{tabular}{l|ccc|ccc}
\toprule
\multirow{2}{*}{Backbones} & \multicolumn{3}{c|}{COME15K-E}     & \multicolumn{3}{c}{COME15K-H}     \\ \cline{2-7} 
                & \multicolumn{1}{c}{AP} & \multicolumn{1}{c}{AP$_{50}$} & AP$_{70}$ & \multicolumn{1}{c}{AP} & \multicolumn{1}{c}{AP$_{50}$} & AP$_{70}$   \\ \hline \hline
      ResNet-50~\cite{he2016deep}     &     58.0     &      75.8        &    65.6     &   50.7        &   70.4      &  57.3   \\
    ResNet-101~\cite{he2016deep}     &  58.5      & 76.4   &   66.0    &  51.5   &  71.8   &   58.0    \\ 
  Swin-T~\cite{liu2021swin}     &    60.0    &  77.9  &   67.5    & 52.6    &   72.9  &   58.6   \\    
\rowcolor[RGB]{235,235,235}
 PVTv2-B2-Li~\cite{wang2022pvt}     &  60.7   &  78.1  &  67.8     &   53.7     &   72.9     &    59.6       \\ 
 \rowcolor[RGB]{235,235,235}
 P2T-Large~\cite{wu2022p2t}     &  \textbf{61.8}   &  \textbf{78.8}  &  \textbf{68.6}     &   \textbf{54.4}     &   \textbf{73.5}     &    \textbf{60.3}      \\  \bottomrule
\end{tabular}
\end{center}
\end{table}

\subsection{Limitations}

\begin{figure}
 \centering
    \begin{overpic}[width=\linewidth]{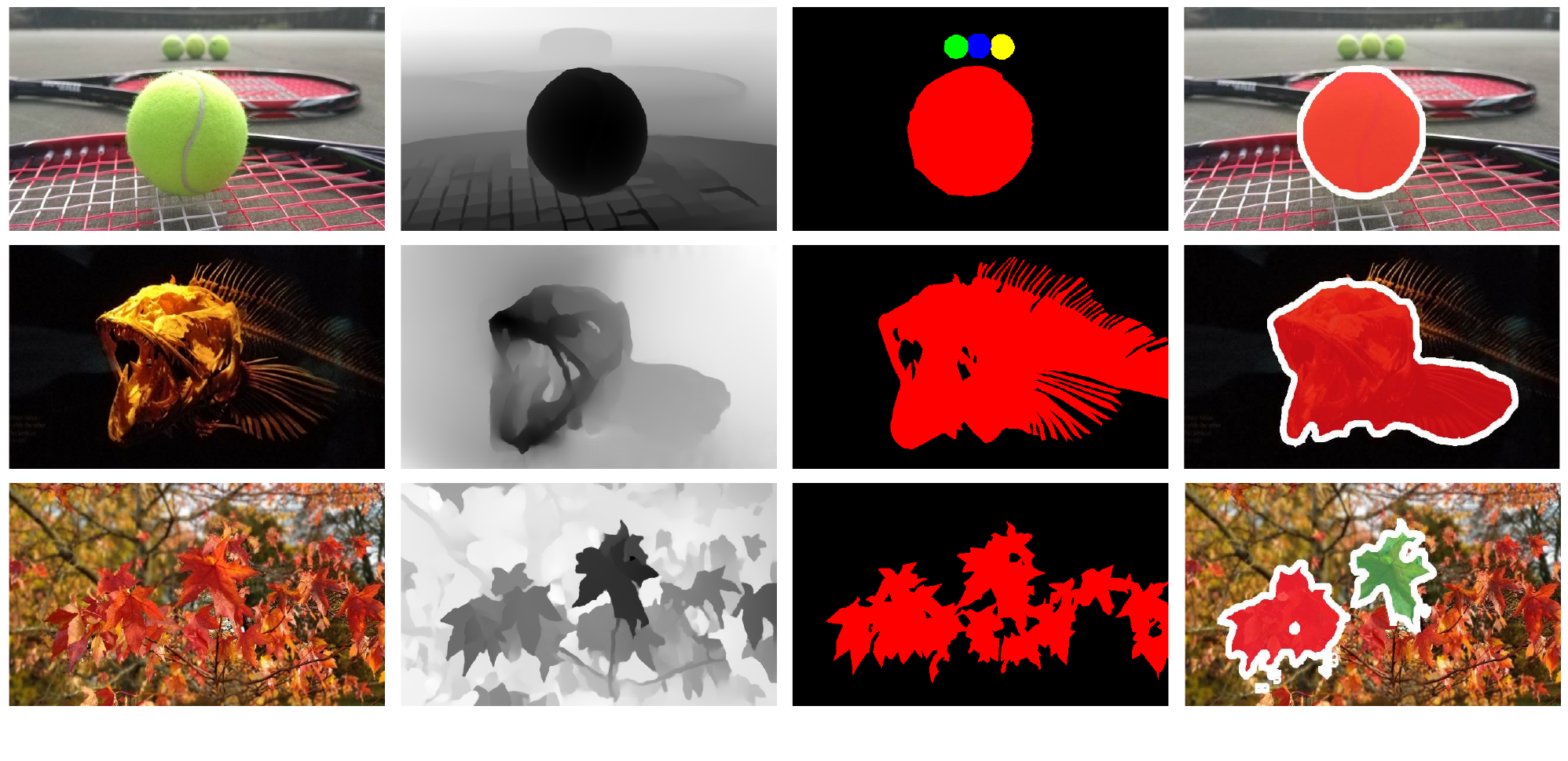}
        \put(9,-0.5){\small{RGB}}
        \put(33,-0.5){\small{Depth}}
        \put(53,-0.5){\small{Instance-GT}}
        \put(83,-0.5){\small{Result}}
    \end{overpic}
 \caption{Typical failure results produced by~\ourmodel.}\label{failure_cases}
\end{figure}
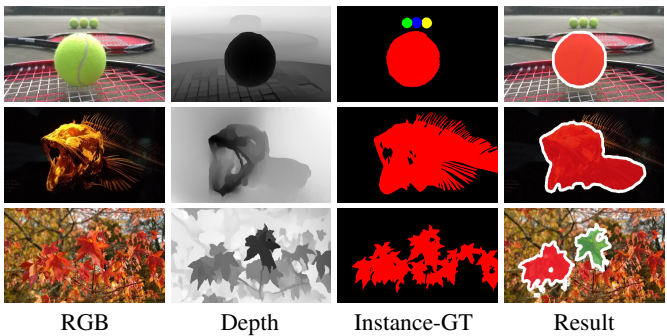

We display several typical failure cases generated by the proposed~\ourmodel~with ResNet-50~\cite{he2016deep} on the COME15K test set in \figref{failure_cases}. 
The first row indicates that~\ourmodel~has a tendency to overlook certain smaller targets when presented with both small and large targets simultaneously.
This could be attributed to the negative feedback given by the depth map, resulting in the generated kernels disregarding the smaller instance embeddings. 
The middle sample illustrates that low contrast and complex object edges can lead to suboptimal segmentation results. 
This problem can be mitigated by introducing edge priors or embedding the operation of highlight edge cues. 
Additionally, if the background of the scene is overly cluttered (see the bottom sample), \ourmodel~has the potential to over-segment salient instances. 
This issue drives us to further improve to handle increasingly complex scenarios, including the ability to better understand and differentiate salient instances from the background in cluttered scenes.


\section{Conclusion}

This paper presents a dual-branch cross-modal calibration model called \ourmodel~for the RGB-D salient instance segmentation task. 
The proposed model consists of two parallel branches: the kernel branch efficiently combines RGB and depth features to generate instance-aware kernels 
using the dynamic interactive kernel (DIK) module, while the mask branch calibrates and integrates multiscale cross-modal feature maps with the weight-sharing fusion (WSF) module to produce a mask feature. 
Additionally, a depth similarity assessment (DSA) module is also included to suppress noisy depth information. 
Numerous experiments conducted demonstrate the benefits of depth information on SIS performance and the superiority of~\ourmodel. 
Furthermore, the authors have contributed a new RGB-D SIS dataset, DSIS, for the comprehensive evaluation of relevant models. 
This task holds great potential for enhancing other multi-modal detection tasks.

\ifCLASSOPTIONcaptionsoff
  \newpage
\fi

\bibliographystyle{IEEEtran}
\bibliography{CalibNet}

%
%




\end{document}